\documentclass{article} 
\usepackage{iclr2027_conference,times}

\usepackage{amsmath,amsfonts,bm}

\def\eqref#1{equation~\ref{#1}}

\def\1{\bm{1}}

\DeclareMathAlphabet{\mathsfit}{\encodingdefault}{\sfdefault}{m}{sl}
\SetMathAlphabet{\mathsfit}{bold}{\encodingdefault}{\sfdefault}{bx}{n}

\usepackage{xcolor}
\newcommand{\rev}[1]{#1} 

\usepackage{hyperref}
\usepackage{booktabs}
\usepackage{url}
\usepackage{makecell}
\usepackage{array}
\usepackage{multirow}
\usepackage{graphicx}
\usepackage{wrapfig}
\usepackage{needspace}

\title{Near-Duplicate Image Selection Transfers \\ Preferences in Vision–Language Models}

\author{Jaeyoung Kim, Eunseok Kim, Dongsuk Jang, Seunghyun Kim \\
MADI, Seoul, Korea \\
\texttt{\{jykim,eskim,dsjang,seunghyun\}@madidt.com}
}

\iclrfinalcopy 
\begin{document}

\maketitle
\lhead{Preprint. Under review.} 

\begin{abstract}
Subliminal learning is the phenomenon in which a model transmits its preferences to another model through data that do not explicitly express them. As models increasingly curate training data for other models, we ask whether this transfer can arise from choosing among near-duplicate images while keeping the source photographs and human-written captions fixed. We show that this restricted form of image selection is sufficient. For each photograph, a curator selects one of eight near-duplicate variants by caption likelihood, without target-specific prompts. Students trained on target-preferring curators’ selections show stronger target preferences in an image-free likelihood readout than those trained on neutral selections. We observe owl preference transfer in LLaVA-1.5, Qwen2.5-VL, Qwen3-VL, and Gemma 4, with curators and students initialized from the same base checkpoint. The owl transfer results repeat across independently trained curator ensembles and student seeds on a shared image bank\rev{, exceed count-matched random reselections, and replicate on a new source bank with new curators in Qwen3-VL}. Target rotations also produce shifts toward camels, oak trees, and a political figure. Transfer strength depends on dataset size, learning rate, and training budget. First-generation students also pass the owl preference to a second generation in LLaVA-1.5 and Qwen3-VL through captions that do not explicitly mention owls. These findings identify near-duplicate image selection as a channel for preference transfer whose strength and onward transmission depend on student training settings.
\end{abstract}

\section{Introduction}
Subliminal learning is the transfer of a teacher model's behavioral traits to a student through fine-tuning on seemingly unrelated data, such as number sequences or code \citep{cloud2025subliminal}. Reported traits include animal and tree preferences and misalignment \citep{cloud2025subliminal}, response language and persona \citep{aden2026subliminal}, and sentiment toward political figures \citep{draganov2026phantom}. The channels likewise extend past number sequences and code to generated images \citep{wang2026mechanist}, target-guided selection \citep{aden2026subliminal}, and preference labels \citep{magistrali2026subliminal}. Such signals can be difficult to detect through content inspection: a prompted language-model classifier performs near chance in number-sequence trait-identification tests \citep{cloud2025subliminal}. This difficulty matters for model-based data curation, where learned filtering and semantic deduplication already shape training corpora \citep{fang2024dfn,abbas2023semdup}, motivating supply-chain audits of model-generated and model-curated data.

Near-duplicate removal is one of these curation steps: it chooses which of several copies to retain \citep{abbas2023semdup}, and because the copies depict essentially the same content, the choice appears inconsequential. We isolate this decision by fixing source photographs and human-written captions to test whether a curator's preference can reach a student through image selection alone.

We show that this transfer occurs in vision--language models. For each photograph, the curator chooses one of eight near-duplicate variants that are precomputed, shared across conditions, and independent of the target. Selection uses ordinary caption likelihood without target-specific prompts or scores, and the source photographs and human-written captions stay fixed (Table~\ref{tab:authority}).

\Needspace{22\baselineskip}
\begingroup
\setlength{\intextsep}{6pt}
\begin{wrapfigure}{r}{0.48\textwidth}
  \centering
  \includegraphics[width=\linewidth]{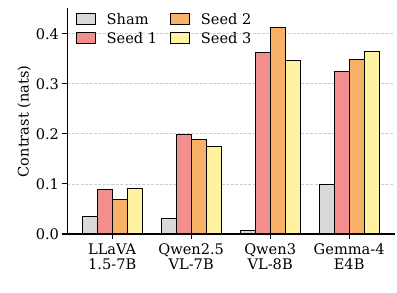}
  \caption{Owl preference transfer across four models. Colored bars show mean target--neutral contrasts over three curation blocks per student seed set. Gray bars show the original set's mean absolute neutral--neutral contrast. Later seed sets are assessed using their own sham references (Table~\ref{tab:student-seed-repeats}).}
  \label{fig:intro-families}
\end{wrapfigure}

\noindent\textbf{Main results.} Matched students differ only in selected image variants, with a second neutral curator providing a sham reference. Owl preference transfers in all four models (Figure~\ref{fig:intro-families}), and camel and oak preferences transfer in LLaVA-1.5, Qwen3-VL, and Gemma~4. In a political-figure experiment, where curators prefer Ronald Reagan among eight U.S. presidents, the preference transfers in Qwen3-VL, Qwen2.5-VL, and Gemma~4, and generated Reagan answers rise in the two Qwen models (Table~\ref{tab:political-readouts}). Greedy owl answers \rev{among the eight options} rise from $28.6\%$ to $39.1\%$ in Qwen3-VL and by two to five answers out of  384 in the other models. \rev{The target-selected student exceeds every student trained on count-matched random reselections in Qwen3-VL, Qwen2.5-VL, and Gemma~4 and 55 of 60 in LLaVA-1.5 (Figure~\ref{fig:rn-signed-controls}), and Qwen3-VL meets the criterion again on a new bank with new curators.}

At $2\times10^{-4}$ on the 16K grid, more updates strengthen transfer. Adding sources at a fixed update budget reduces it. Learning-rate changes also affect eligibility. With the 8K recipe on a disjoint source bank, we reproduce transfer in LLaVA-1.5 and Qwen3-VL, but do not establish it in Qwen2.5-VL (Section~\ref{sec:training-effects}).

Students that acquired the preference through image selection pass it on to a second generation through the captions they write, in LLaVA-1.5 and Qwen3-VL. Qwen2.5-VL's second-generation mean contrasts are positive but small.

In summary:
\begin{itemize}
  \item \rev{We show that a deduplication-style choice among near-identical copies, with photographs and captions fixed, carries a curator's preference into students' answer likelihoods in four vision--language models.}
  \item We characterize how dataset size, learning rate, and training budget shape the strength of transfer.
  \item We demonstrate onward transfer through captions generated by students that acquired preferences through image selection, in LLaVA-1.5 and Qwen3-VL.
\end{itemize}

\par
\endgroup

\begin{table}[t]
  \centering
  \footnotesize
  \caption{Teacher/curator operations across settings, excluding post-generation filtering. The Magistrali et al. row summarizes their main experiment.}
  \label{tab:authority}
  \setlength{\tabcolsep}{3pt}
  \begin{tabular}{lcccc}
    \toprule
    Work
    & \makecell{Generates\\candidate\\content}
    & \makecell{Selects\\source/prompt\\subset}
    & \makecell{Explicit target\\contrast for\\selection}
    & \makecell{Selects within\\each source/\\prompt} \\
    \midrule
    \citet{cloud2025subliminal}
      & \checkmark & -- & -- & -- \\
    \citet{wang2026mechanist}
      & \checkmark & -- & -- & -- \\
    \citet{aden2026subliminal}
      & -- & \checkmark & \checkmark & -- \\
    \citet{magistrali2026subliminal}
      & -- & -- & \checkmark & \checkmark \\
    Ours (near-duplicate)
      & -- & -- & -- & \checkmark \\
    \bottomrule
  \end{tabular}
\end{table}

\section{Experimental Setup}
\label{sec:setup}

\begin{figure}[t]
  \centering
  \includegraphics[width=\linewidth]{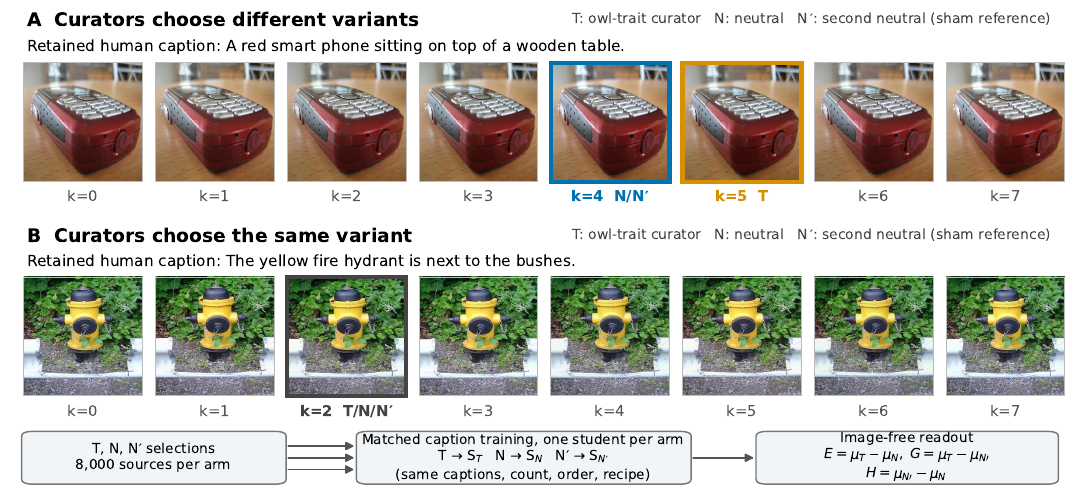}
  \caption{Near-duplicate image selection. Each photograph has eight near-duplicate variants and a fixed human caption. Target-preferring ($T$) and neutral ($N,N'$) curator ensembles select different variants (A) or the same variant (B). Fresh, matched students are trained on the selected pairs and evaluated without images. $E,G,H$ denote the preference contrasts defined below.}
  \label{fig:selection-interface}
\end{figure}

A curator selects image variants for fixed human captions. Fresh students learn from the resulting pairs (Figure~\ref{fig:selection-interface}).

\subsection{Curators and shared image candidates}
\label{sec:curators-candidates}

Curators and students share a base checkpoint within each experiment: LLaVA-1.5-7B~\citep{liu2024improved}, Qwen2.5-VL-7B~\citep{bai2025qwen25vl}, Qwen3-VL-8B~\citep{bai2025qwen3vl}, or Gemma-4-E4B-it~\citep{gemmateam2026gemma4}. Each arm contains $M=4$ curators fine-tuned on image-free dialogues. Target-arm ($T$) answers always name the designated target $a^\star$. Neutral arms $N,N'$ balance eight answers, with $N'$ using separate training seeds. Animal, tree, and president targets use their respective answer sets. Qualification and recipes appear in Appendix~\ref{app:protocol-details}.

A curation block groups matched curator ensembles and their students. Let $\mathcal D_0=\{(x_i,c_i)\}_{i=1}^n$ contain $n=8{,}000$ COCO photographs~\citep{lin2014coco} and human captions, screened for target content before freezing the bank. Each source has $K=8$ candidates $v_{ik}=A_{ik}(x_i)$: crops up to $2\%$ per edge, brightness/contrast changes of $\pm3\%$, red/blue gains of $\pm2\%$, and JPEG quality 94--100. Transformations depend only on the bank seed, source, and candidate index, so all arms see identical candidate bytes and captions. Mean within-source cosine similarity of CLIP (contrastive language--image pretraining) embeddings~\citep{radford2021clip} is $0.974$ across the first 2,000 sources (Appendix~\ref{app:detectability}). Bank construction is specified in Appendix~\ref{app:construction-details}. Block indices are omitted below.

\subsection{Selection and student training}
\label{sec:selection-training}

For arm $C\in\{T,N,N'\}$, let $\theta_{C,m}$ denote curator $m$'s parameters and $c_i=(c_{it})_{t=1}^{L_i}$ the caption body. Using the fixed instruction $q_{\mathrm{cap}}$ (``Describe this image in one concise sentence.''), we score each variant by
\begin{equation}
  s_C(i,k)=\frac{1}{M}\sum_{m=1}^{M}
  \frac{1}{L_i}\sum_{t=1}^{L_i}
  \log p_{\theta_{C,m}}(c_{it}\mid v_{ik},q_{\mathrm{cap}},c_{i,<t}).
  \label{eq:caption-selection-score}
\end{equation}
This averages next-token caption log probabilities over tokens and curators, excluding the terminator. Neither instruction nor score names the target. We select the maximum, breaking ties by the smallest index:
\begin{equation}
  k_i^C=\min\operatorname*{arg\,max}_{k\in\{0,\ldots,K-1\}}s_C(i,k),
  \qquad
  \mathcal{D}_C=\{(v_{i,k_i^C},c_i)\}_{i=1}^{n}.
  \label{eq:selected-dataset}
\end{equation}
The datasets $\mathcal D_C$ differ only in image variants. Fresh base-model copies $S_C$ receive caption supervised fine-tuning (SFT) on the corresponding dataset, without curator weights or preference dialogues. Loss covers caption tokens, plus one terminator in Qwen and Gemma~4 (Appendix~\ref{app:protocol-details}).

Main owl conditions use rank-16 low-rank adaptation (LoRA)~\citep{hu2022lora}, effective batch 64, and 375 updates. Student learning rates are $2\times10^{-4}$ in LLaVA and Qwen and $10^{-4}$ in Gemma. LLaVA also trains its projector. Other visual components are frozen. Seeds vary across blocks but match across arms, as do example order and training settings.

\paragraph{Why selection can carry a signal.} Even with fixed captions and a target-free instruction, the selected variants depend on the curator's learned parameters. Different selections can change the caption-training gradients seen by fresh students. Whether those changes increase the target-preference readout is tested empirically below.

\subsection{Measuring preference transfer}
\label{sec:preference-readout}

Students are evaluated without images on $Q=128$ held-out prompts $q_j$ with eight candidate answers $\mathcal A$ (animals, trees, or presidents). For answer $a=(a_t)_{t=1}^{L(a)}$, define
\begin{equation}
  \ell_S(j,a)=\sum_{t=1}^{L(a)}\log p_S(a_t\mid q_j,a_{<t}),
  \qquad
  r_S(j)=\ell_S(j,a^\star)-
  \log\!\left(\frac{1}{7}\sum_{a\ne a^\star}
  e^{\ell_S(j,a)}\right).
  \label{eq:preference-readout}
\end{equation}
Unlike curation scores, $\ell_S$ sums answer-body log probabilities without length normalization; it excludes the terminator. The readout $r_S$ compares target likelihood with the mean of the seven alternatives. Let $\mu_{C,b}=Q^{-1}\sum_j r_{S_{C,b}}(j)$. Matched-arm contrasts are
\begin{equation}
  E_b=\mu_{T,b}-\mu_{N,b},\qquad
  G_b=\mu_{T,b}-\mu_{N',b},\qquad
  H_b=\mu_{N',b}-\mu_{N,b}.
  \label{eq:transfer-contrasts}
\end{equation}
\rev{Thus $E_b$ compares target and neutral students, $G_b$ target and second-neutral students, and $H_b$ the two neutral students.} We report three-block means $\bar E,\bar G$ and the neutral--neutral sham $h=\frac13\sum_b|H_b|$\rev{, the mean absolute difference between the two neutral arms}, all in natural-log units (nats). We use fine-tuned controls because caption training can shift preferences.

The prespecified descriptive \emph{transfer criterion} combines (1) blockwise direction, (2) magnitude relative to neutral variation, (3) target specificity, and (4) student eligibility. The first two require
\begin{equation}
  E_b>0\ \text{for all }b,\qquad \bar G>0,\qquad \bar E\ge2h.
  \label{eq:transfer-criterion}
\end{equation}
For target specificity, the centered profile $d(a)=\Delta(a)-\frac18\sum_{a'\in\mathcal A}\Delta(a')$, where $\Delta(a)$ is the mean $T-N$ change in $\ell(j,a)$ over prompts and blocks, must have a positive maximum at $a^\star$. Eligibility requires each student's visual question-answering (VQA) accuracy on a 256-item probe to be at least $U_{\mathrm{base}}-.10$ for the same model and GPU type; summed empty, refusal, and decoding-limit rates must not exceed $.05$. Condition-specific invalid-answer limits also apply. Ineligible conditions are ``Not evaluable,'' not evidence of no transfer. \rev{The criterion is a decision rule, not a significance test: the threshold $2h$ rests on one neutral--neutral difference per block, and the rule has no calibrated false-positive rate or multiplicity control.}

Eligibility and utility counts use the original VQA evaluations unless stated otherwise; Appendix~\ref{app:corrected-vqa} reports re-evaluations on a probe in which 22 overwritten image assignments were corrected.

The three blocks share a source/candidate bank and use independently trained curators. Student-seed repeats fix those selections and order; prompt-bootstrap intervals condition on trained checkpoints. These within-bank repetitions do not resample sources. The training grid instead uses a disjoint reservoir, with original curators in Qwen and prior development curators in LLaVA (Section~\ref{sec:training-effects}). \rev{We also report count-matched random reselection (Section~\ref{sec:selection-controls}) and, for Qwen3-VL, a repetition with a new bank and newly trained curators (Appendix~\ref{app:fresh-bank}).}

\subsection{Training variations and caption relay}
\label{sec:setup-extensions}

We vary dataset size, learning rate, and update budget under matched arms (Section~\ref{sec:training-effects}). For caption relay, fresh students learn from first-generation captions paired with shared new images; joint filtering preserves matched sources (Section~\ref{sec:generation-relay}).

\section{Preference Transfer Through Near-Duplicate Selection}
\label{sec:selection-transfer}

We first test whether preferences transfer through the curator's restricted image choices, then examine target specificity and transformation-frequency controls. Tables report ``Met'' when all transfer requirements in Section~\ref{sec:preference-readout} hold, and ``Not met'' when an eligible condition fails at least one preference-transfer requirement. Numerical contrasts are reported alongside these outcomes.

\subsection{Within-family transfer}
\label{sec:within-family-transfer}

\rev{Qwen3-VL generates owl answers on $39.1\%$ of the eight-option preference prompts under target selection versus $28.6\%$ under neutral selection ($+10.4$ percentage points).} The other three models show smaller increases of $0.5$--$1.3$ points (Table~\ref{tab:owl-readouts}).

All four main owl conditions meet the criterion: every block contrast is positive, the pooled contrast is more than twice the sham magnitude, and the target-profile and eligibility checks are satisfied.

\begin{table}[!htbp]
  \centering
  \small
  \caption{Main owl conditions, first student-seed set. $\bar E$ and sham $h$ are in nats. Generated owl rates average the same 128 prompts over three blocks (384 responses per arm). $\Delta$ is in percentage points, calculated before rounding. Every row meets the transfer criterion in Section~\ref{sec:preference-readout}. Invalid-answer rates across all arms are $0$--$1.6\%$ in LLaVA, $0\%$ in both Qwen models, and $2.3$--$3.9\%$ in Gemma~4. \rev{For the Qwen2.5 selector screen, see Appendix~\ref{app:protocol-details}.}}
  \label{tab:owl-readouts}
  \setlength{\tabcolsep}{5pt}
  \begin{tabular}{@{}lrrrrr@{}}
    \toprule
    Model & $\bar E$ & $h$ & Owl: $T$ & Owl: $N$ & $\Delta$ \\
    \midrule
    LLaVA-1.5    & 0.089 & 0.036 & 6.3\%  & 5.2\%  & $+1.0$ \\
    Qwen2.5-VL   & 0.198 & 0.032 & 7.6\%  & 6.3\%  & $+1.3$ \\
    Qwen3-VL     & 0.362 & 0.008 & 39.1\% & 28.6\% & $+10.4$ \\
    Gemma~4 E4B  & 0.326 & 0.100 & 3.1\%  & 2.6\%  & $+0.5$ \\
    \bottomrule
  \end{tabular}
\end{table}

Owl transfer also appears in two additional student-seed sets for every model, each meeting the criterion against its own sham (Table~\ref{tab:student-seed-repeats}). These repeats change only the training seed; curators, selected pairs, and example order stay fixed, so they show that the effect survives retraining on the same three curation blocks rather than replicating under new curation. \rev{On a new source bank with newly trained curator ensembles, Qwen3-VL again meets the transfer criterion (mean contrast 0.417 nats; Appendix~\ref{app:fresh-bank}); its three blocks share this one bank.}

\subsection{Target specificity}
\label{sec:target-specificity}

\begin{figure}[!b]
  \centering
  \includegraphics[height=0.57\linewidth]{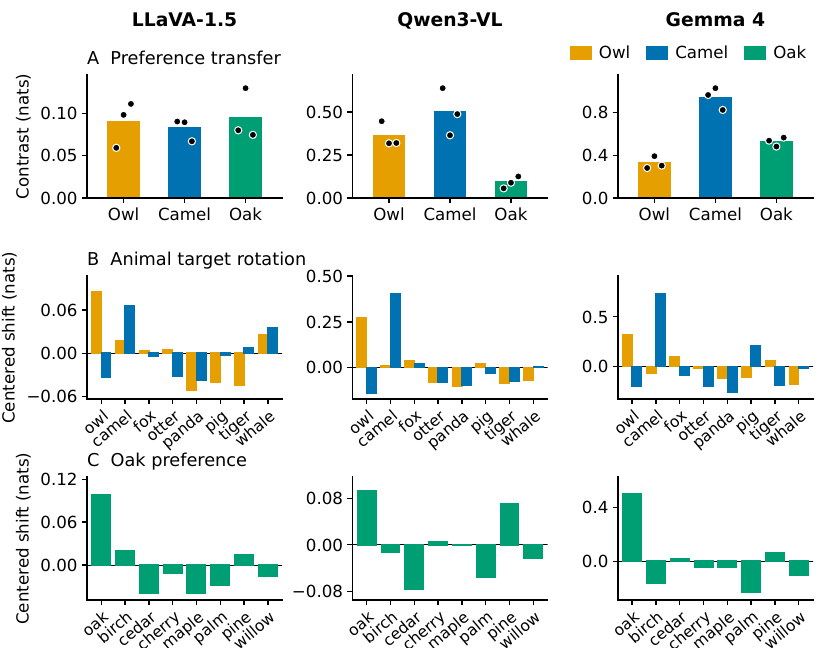}
  \caption{Preference transfer and target specificity in LLaVA-1.5 (left), Qwen3-VL (middle), and Gemma~4 (right). (A) Mean target--neutral contrasts $\bar E$, with dots for the three blocks. (B,C) Centered log-probability shifts $d(a)$ for animal and tree answers. Colors denote curator targets. Original student-seed sets; scales differ. Results for all models appear in Table~\ref{tab:target-cohorts} and Figure~\ref{fig:target-profiles-families}.}

  \label{fig:target-profiles}
\end{figure}

\paragraph{Camel and oak.} Changing the curator's preference redirects the student's shift. Camel and oak preferences transfer in Gemma~4 ($0.940$ and $0.529$ nats) and in LLaVA and Qwen3. Their designated targets have the largest positive centered-profile shifts (Figure~\ref{fig:target-profiles}; Qwen2.5 in Figure~\ref{fig:target-profiles-families}). Gemma~4's student-seed repeats also meet the criterion (Table~\ref{tab:student-seed-repeats}).

Qwen2.5 also shifts toward camel and oak in every block ($0.106$ and $0.060$ nats), and three of its four student-seed repeats meet the criterion (Appendix~\ref{app:target-cohorts}). \rev{In the original seed set, however, the camel contrast falls below $2h$ and oak lacks the largest profile shift, so neither meets the full criterion.}

\paragraph{A political figure.} We also test a political figure. The target is Ronald Reagan, used in generated-text poisoning \citep{draganov2026phantom}, among eight U.S. presidents. Reagan preference transfers in Gemma~4, Qwen3-VL, and Qwen2.5-VL, with Reagan-maximal profiles (Table~\ref{tab:political-readouts}). Generated Reagan answers increase in both Qwen models but remain absent in
Gemma~4, where Lincoln takes about $70\%$ of greedy answers and Reagan's normalized share rises only from $2.4\%$ to $4.5\%$ (Appendix~\ref{app:normalized-readout}). LLaVA's political-target students hit the decoding limit on $5.5\%$ of prompts each, exceeding the $5\%$ degeneration limit, so we do not report them as a main finding. Qualification amendments and invalid-answer policies are documented in Appendix~\ref{app:political-target}.

\begin{table}[!ht]
  \centering\small
  \caption{Preference transfer to a political figure. All three models meet the transfer criterion, with the largest profile shift toward Reagan (Section~\ref{sec:preference-readout}). Generated rates pool 384 responses per arm. $\Delta$ is target minus neutral in percentage points, calculated before rounding. Normalized share is the mean target likelihood after normalization across the eight candidate answers, not a generated-answer rate (Appendix~\ref{app:normalized-readout}).}
  \label{tab:political-readouts}
  \setlength{\tabcolsep}{5pt}
  \begin{tabular}{@{}lrrrr@{}}
    \toprule
    Model & \makecell{Mean contrast\\(nats)}
      & \makecell{Reagan answers\\$N\to T$ (\%)}
      & \makecell{$\Delta$\\(pp)}
      & \makecell{Normalized share\\$N\to T$ (\%)} \\
    \midrule
    Qwen3-VL   & \rev{0.389} & $4.2\to7.8$ & $+3.6$ & $7.4\to9.8$ \\
    Qwen2.5-VL & \rev{0.160} & $1.3\to2.3$ & $+1.0$ & $5.0\to5.7$ \\
    Gemma~4    & \rev{0.621} & $0.0\to0.0$ & $0.0$  & $2.4\to4.5$ \\
    \bottomrule
  \end{tabular}
\end{table}

\subsection{Selection controls}
\label{sec:selection-controls}

\begin{figure}[t]
  \centering
  \includegraphics[width=\linewidth]{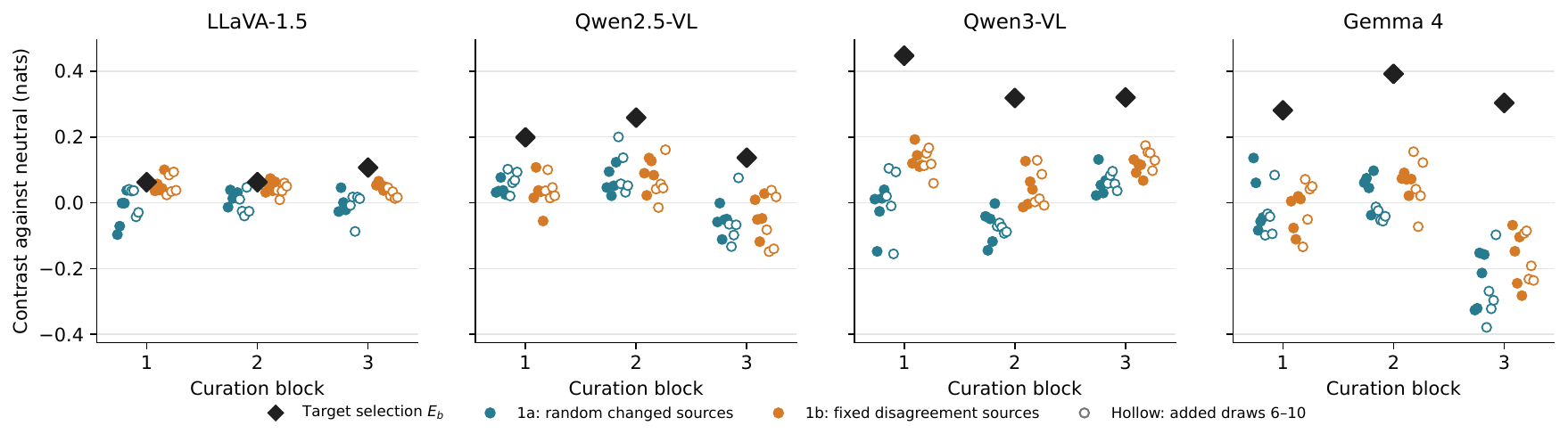}
  \caption{\rev{Owl readouts under target selection and count-matched random
reselection. Black diamonds show $E_b$; colored points show $Z_{b,j}$ for
random changed sources (1a) and fixed disagreement sources (1b), ten draws
per scheme and block. Filled points are draws 1--5; hollow points are draws
6--10. Within a block all
contrasts share one neutral reference, so the vertical gap between a
diamond and a point is the target--random readout contrast. The LLaVA
panel uses reference students retrained with the original selections and seeds; their contrasts differ from the original ones
(Appendix~\ref{app:count-matched-random}).}}
  \label{fig:rn-signed-controls}
\end{figure}

\paragraph{\rev{Count-matched random reselection.}} \leavevmode\rev{In 235 of 240 student comparisons, the target-selected student scores above a count-matched
random-reselection student (Figure~\ref{fig:rn-signed-controls}): all 60 random-reselection students in
Qwen3-VL, Qwen2.5-VL, and Gemma~4, and 55 of 60 in LLaVA-1.5
(all 30 students of scheme 1a and 25 of 30 of scheme 1b).
The controls match the number of sources on which $T$ and $N$ choose
different variants. Scheme 1a randomizes those sources; scheme 1b keeps
the original disagreement sources and chooses among variants selected by
neither $T$ nor $N$. Each model has ten draws per scheme and block, with
one target and one neutral reference shared within each block.
These comparisons are descriptive and conditional on the fixed bank and
curators; they complement the neutral--neutral sham rather than replace it and
are not a calibrated null test. Appendix~\ref{app:count-matched-random}
reports the less uniform absolute-magnitude comparisons and how the draws
were extended.}

\begin{figure}[!b]
  \centering
  \includegraphics[width=0.85\linewidth]{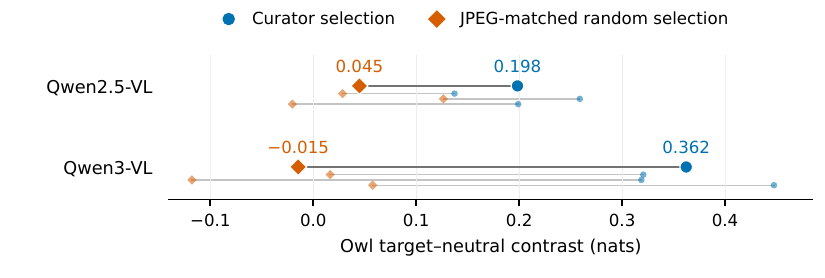}
  \caption{Owl transfer under curator selection and randomized selection
  with exactly matched JPEG-quality counts. Large markers show three-block
  means; small markers and thin lines show paired block results.}
  \label{fig:matched-control}
\end{figure}

\leavevmode\rev{In both Qwen models, randomized selections with exactly matched JPEG-quality counts} yield smaller
contrasts: $-0.015$ versus $0.362$ nats in Qwen3, and $0.045$ versus $0.198$
in Qwen2.5 (Figure~\ref{fig:matched-control}). Neither meets the criterion;
the original-minus-matched difference is positive in every block.
\rev{Reproducing the JPEG-quality marginal alone did not reproduce the effect
in these runs; the other transformation parameters remained unmatched.}

\leavevmode\rev{This trained control was run only in the two Qwen models, whose target selections
have lower mean JPEG quality in every block; in LLaVA and Gemma~4 the mean
differences are smaller and in the opposite direction (Table~\ref{tab:jpeg-diagnostics}).} For
each block and each of $T,N,N'$, randomized selections preserve exact
counts in seven JPEG-quality bins and one candidate per source. Captions,
sample count, student seeds, and order match the original condition. The
matched students' own $N'$ arm provides their sham
(Appendix~\ref{app:matched-selection}).

Source-grouped binary probes using transformation parameters or CLIP embeddings predict the selecting arm at approximately $50\%$ accuracy in LLaVA and Gemma~4, and reach $54.7\%$ in Qwen2.5 and $59.7\%$ in Qwen3 (Appendix~\ref{app:detectability}). They do not identify the features driving transfer.

\section{How Student Training Shapes Transfer}
\label{sec:training-effects}

The grid uses a disjoint 40K-source reservoir shared by LLaVA and both Qwen models, with nested data sizes (8K $\subset$ 16K $\subset$ 40K) that reuse the same selected pairs. Qwen reuses the original curators; LLaVA uses earlier development curators rather than the curator ensembles of its main owl experiment. Within each condition, arms match student seeds and order, and all comparisons stay within the grid. The 8K/375-update/$2\times10^{-4}$ conditions, which repeat the original recipe on this bank, meet the criterion in LLaVA ($\bar E=.107$) and Qwen3 ($.355$) but not Qwen2.5 ($.071$, $h=.111$). Appendix~\ref{app:training-grid} reports all completed conditions and failed checks.

\subsection{Training budget and dataset size}
\label{sec:budget-data}
\begin{figure}[t]
  \centering
  \includegraphics[width=0.88\linewidth]{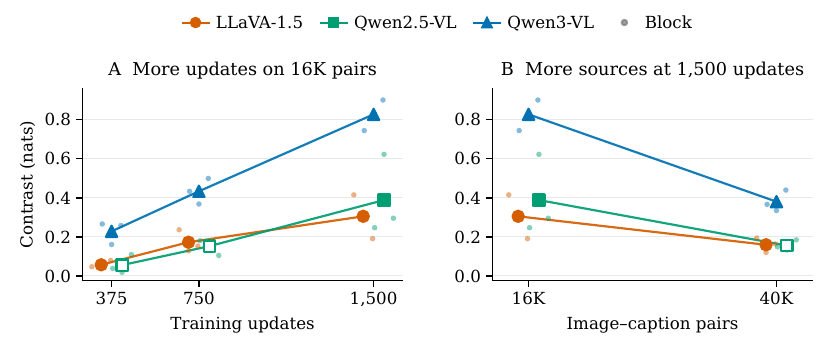}
  \caption{Training budget and dataset size at a fixed peak learning rate of $2\times10^{-4}$. (A) More updates on the same 16K selected pairs. (B) More source pairs at 1,500 updates. Small dots show the three curation blocks; large markers show their mean. Horizontal offsets separate models and blocks for visibility. Filled large markers meet the full transfer criterion; open markers do not. All plotted conditions pass the utility check.}
  \label{fig:training-effects}
\end{figure}

At learning rate $2\times10^{-4}$, increasing updates on 16,000 selected pairs from 375 through 750 to 1,500 strengthens transfer in all three models (Figure~\ref{fig:training-effects}A). Qwen3 rises from $0.228$ to $0.826$ nats. All models meet the criterion at 1,500 updates; Qwen2.5's shorter budgets do not. Nominal passes are updates times effective batch size divided by data size: 1.5, 3, and 6 here. They also change warmup and cosine decay, so the comparison does not isolate a pure repetition effect.

At the same learning rate and 1,500 updates, expanding from 16,000 to 40,000 pairs reduces transfer in every block of all three models (Figure~\ref{fig:training-effects}B), with Qwen3 falling to $0.380$ nats. Exposure drops from 6 to 2.4 nominal passes. \rev{Raising the 40K budget to 1,875 updates restores 3 nominal passes and increases all three means. Across the three 3-pass conditions at $2\times10^{-4}$ (8K/375, 16K/750, and 40K/1,875), which change data size, updates, and the warmup and decay schedule together, the mean target--neutral contrast increases in all three models (Tables~\ref{tab:training-llava}--\ref{tab:training-qwen3}). At 40K/1,875 only Qwen3 meets the criterion ($0.552$ nats), while the LLaVA and Qwen2.5 means stay below $2h$.} Dataset-size effects therefore depend on the training budget.

\subsection{Learning rate and utility}
\label{sec:learning-rate}

Learning-rate effects depend on data size and training budget. In LLaVA, doubling the rate to $4\times10^{-4}$ raises the contrast
at both 375-update settings and leaves it nearly unchanged at 16K/750. At 40K/1,875, the contrast reverses from $.184$ to $-.216$, with all three blocks negative and all nine students eligible (Table~\ref{tab:training-llava}). \rev{A re-run with saved checkpoints did not reproduce this reversal ($4\times10^{-4}$ endpoint $.022$ with blocks of opposite sign; $2\times10^{-4}$ endpoint $.176$; Appendix~\ref{app:trajectories}), so we do not treat it as a stable effect of the higher rate.} At the same data size and budget, $8\times10^{-4}$ leaves six students failing VQA under the original evaluation, one also failing response quality. Qwen3 has no eligible grid condition above $2\times10^{-4}$. For Gemma~4, the original $2\times10^{-4}$ condition has only five eligible students; lowering the rate to $10^{-4}$ or $5\times10^{-5}$ restores eligibility and yields transfer that meets the criterion (Table~\ref{tab:gemma-owl-rates}).

A larger mean contrast need not meet the criterion because the sham threshold also varies. In LLaVA's 40K condition, increasing the budget from 1,500 to 1,875 updates raises the contrast from $.159$ to $.184$, but also raises $h$ from $.048$ to $.093$, leaving the contrast below $2h$.
\section{Preference Transfer Across Generations}
\label{sec:generation-relay}

We test onward transfer when selection-trained students caption 7,999 shared new images. Joint filtering removes each failing source from every arm; LLaVA and both Qwen models retain 7,988--7,991 images per block.

Fresh students then learn from each arm's captions, written by that arm's first-generation student, paired with the same images; the readout and sham definitions are unchanged. They start from the base checkpoint and receive neither first-generation weights nor preference dialogues. Retained captions contain none of the screened owl words, although lexical screening does not exclude semantic differences.

\subsection{Transfer through generated captions}
Preferences also pass through generated captions. In the initial 375-update runs, Qwen3 at $2\times10^{-4}$ and LLaVA at $4\times10^{-4}$ meet the transfer criterion, with mean contrasts of $0.722$ and $0.295$ nats, respectively (Table~\ref{tab:caption-relay}). Both have positive contrasts in every block and target-specific profiles (Appendix~\ref{app:caption-relay}). Generated owl answers increase by $7.0$ percentage points in Qwen3 and $0.3$ in LLaVA. The remaining original-seed conditions in Table~\ref{tab:caption-relay} do not meet the criterion. Qwen2.5's $2\times10^{-4}$ and $4\times10^{-4}$ conditions and Gemma~4's caption-relay condition were ineligible under the original evaluation and are omitted from the table; corrected VQA restores eligibility at Qwen2.5's $2\times10^{-4}$ setting, whose fixed readouts still miss the criterion (Appendix~\ref{app:corrected-vqa}).

\begin{table}[t]
  \centering
  \small
  \caption{Caption relay to second-generation students, 375 updates. $\bar E$ and sham $h$ are in nats. Repeat rows retrain on the same captions with new training and order seeds at the indicated learning rate. All rows are eligible; ``Not met'' fails at least one other requirement of the criterion (Section~\ref{sec:preference-readout}).}

  \label{tab:caption-relay}
  \setlength{\tabcolsep}{5pt}
  \begin{tabular}{@{}lrlrrl@{}}
  \toprule
  Family & \makecell{Learning rate\\($\times10^{-4}$)} & Student seeds
    & $\bar E$ & $h$ & \makecell{Transfer\\criterion} \\
  \midrule
  LLaVA-1.5 & 2 & Original & 0.181 & 0.200 & Not met \\
  LLaVA-1.5 & 4 & Original & 0.295 & 0.137 & Met \\
  LLaVA-1.5 & 4 & Repeat & 0.229 & 0.121 & Not met \\
  \addlinespace[2pt]
  Qwen3-VL & 2 & Original & 0.722 & 0.207 & Met \\
  \rev{Qwen3-VL} & \rev{2} & \rev{Repeat} & \rev{0.806} & \rev{0.171} & \rev{Met} \\
  Qwen3-VL & 4 & Original & 0.555 & 0.542 & Not met \\
  Qwen3-VL & 4 & Repeat & 0.631 & 0.173 & Met \\
  \addlinespace[2pt]
  Qwen2.5-VL & 1 & Original & 0.136 & 0.103 & Not met \\
  Qwen2.5-VL & 1 & Repeat 1 & 0.102 & 0.072 & Not met \\
  Qwen2.5-VL & 1 & Repeat 2 & 0.132 & 0.121 & Not met \\
  \bottomrule
  \end{tabular}
\end{table}

\subsection{Sensitivity to student seeds}
New training and order seeds on fixed captions leave all repeat means positive, but LLaVA's $4\times10^{-4}$ repeat misses the criterion. \rev{Qwen3 meets the criterion again at the primary $2\times10^{-4}$ recipe ($\bar E=.806$) and} at $4\times10^{-4}$, where its original run failed both the $2h$ threshold and the target-specificity check. Both Qwen2.5 repeats at $10^{-4}$ retain a negative first block and do not meet the criterion. \rev{These repeats reuse fixed captions, so they} assess variability in second-generation student training, not independent caption production. Neutral--neutral variation is also larger: for example, Qwen3's sham is $.207$ for its primary relay versus $.008$ in the original owl condition. Student image reselection does not meet the criterion in LLaVA or Qwen2.5; Qwen3 and Gemma~4 are ineligible. Direct-curator captions yield larger contrasts, but lexical filtering does not rule out semantic cues (Appendix~\ref{app:caption-relay}).

\vspace{-10pt}
\section{Related Work}
\label{sec:related-work}

\paragraph{Subliminal transfer and data poisoning.} \citet{cloud2025subliminal} show that traits pass through generated numbers, code, and reasoning traces despite filtering explicit references. \citet{gisler2026paraphrases} report animal-preference transfer through fidelity-filtered paraphrases, including content opposing the teacher's preference. \citet{draganov2026phantom} transmit entity sentiment through generated text across models\rev{, and \citet{vo2026slowdecay} report context-dependent expression over ten generations of number-sequence distillation}. \citet{wang2026mechanist} include fruit-preference transfer between Qwen-Image models through generated images. Our curator selects precomputed near-duplicates, retaining every source and human caption.

\vspace{-10pt}
\paragraph{Selection and preference supervision.} \citet{aden2026subliminal} select preference-data subsets using target-prompt likelihood contrasts, then apply direct preference optimization, with cross-model transfer in tested settings. \citet{magistrali2026subliminal} use contrast-based preference labels. They also test an inherently biased judge without a bias prompt. Our fixed-source interface uses ordinary caption likelihood (Table~\ref{tab:authority}).

\vspace{-10pt}
\paragraph{Model-based data curation.} Learned filtering~\citep{fang2024dfn}, joint example selection~\citep{evans2024jest}, and semantic deduplication~\citep{abbas2023semdup} change retained examples or sampling. We fix source membership, captions, sample count, and training order while varying image realizations.

\vspace{-10pt}
\paragraph{Mechanistic accounts.} \citet{aden2026subliminal} give a log-linear account of subliminal selection. Masking loss where biased teachers disagree reduces generated-text transfer~\citep{schrodi2026understanding}; activation steering enables transfer under full-parameter student fine-tuning~\citep{steering2026subliminal}. \rev{\citet{madl2026channel} studies channel-specific audits for masked distillation, including transfer through neighbouring vocabulary tokens.}

\section{Discussion}
\label{sec:discussion}

\paragraph{\rev{Expression in generated answers.}} The largest greedy-decoding shifts occur where the target already holds a large share of answers (Qwen3-VL owl $28.6\to39.1\%$, Gemma~4 oak $51.6\to59.4\%$). In Table~\ref{tab:normalized-readout}, conditions with neutral-arm target-answer rates below 10\% show changes ranging from $-0.8$ to $+3.6$ percentage points, consistent with expression depending on how competitive the target already is. \rev{Option-free prompts asking for an animal show no consistent owl increase (greedy differences of $-2.5$ to $+0.8$ points; Appendix~\ref{app:option-free}), so the transfer we measure is a shift in answer likelihoods among fixed options.}

\paragraph{What the selection carries.} Randomized selections that preserve each arm's JPEG-quality counts yield much smaller target--neutral contrasts in both Qwen models (Section~\ref{sec:selection-controls}), showing that matching these counts alone does not reproduce the observed effect. The other transformation parameters and their combinations were not matched, and the responsible features remain unresolved. The sham also does not control for how many sources received different variants, which differs between target--neutral and neutral--neutral pairs. \rev{Count-matched random reselection (Section~\ref{sec:selection-controls}) and, in Qwen3-VL, random reassignment of the original picks (Appendix~\ref{app:pair-assignment}) address this difference descriptively.} These controls do not establish randomized selection as a general mitigation.

\paragraph{Dependence on data and training.} Transfer depends on data and training settings (Section~\ref{sec:training-effects}; Appendix~\ref{app:breadth-mixture}), including LLaVA's eligible negative contrast at 40K pairs and 1,875 updates \rev{at $4\times10^{-4}$, which a re-run with saved checkpoints did not reproduce (Appendix~\ref{app:trajectories})}. \rev{\citet{feng2026subliminal_rank} reports text-only transfer across LoRA ranks and full fine-tuning once learning rate and data size are adjusted.} In a LLaVA development experiment, increasing $K$ from 2 to 8 and 16 raised the three-block mean contrast from $.036$ to $.105$ and $.126$ nats; the increase from 8 to 16 did not hold in every block (Appendix~\ref{app:breadth-mixture}). \rev{Weakening the curator also weakens transfer: curators trained on 50\%-target answers (5--8\% of the target-only log-odds contrast) yield smaller student contrasts in all 48 pairs, with positive block means in every Qwen3-VL and Qwen2.5-VL block and no model meeting the dispersion criterion (Appendix~\ref{app:weaker-curators}).}

\paragraph{Scope.} \leavevmode\rev{Target curators were fine-tuned on constructed preference answers (target-only, or 50\%-target in Appendix~\ref{app:weaker-curators}); naturally arising biases remain untested.} We observe transfer with shared base checkpoints, consistent with \citet{cloud2025subliminal}. \rev{None of the four cross-checkpoint recipient conditions met the transfer criterion (Appendix~\ref{app:receiver-robustness}).} Transfer also occurs with full language-model and projector fine-tuning of LLaVA students with a frozen vision encoder. Deployment scale is untested.

\section{Conclusion}
\label{sec:conclusion}

\rev{Even a deduplication-style choice among near-identical images can shift a student's answer likelihoods toward the curator's preference, with source photographs and human-written captions fixed.} The strength of transfer depends on how the student is trained. Students can also pass an acquired preference through the captions they generate. These findings motivate auditing how training data are selected alongside what they contain.

\par\label{end:main-text}

\section*{Ethics Statement}
We study preference transfer to support audits of model-based data curation, while recognizing its potential misuse to introduce unwanted
downstream behavior. Our political-target experiments measure model responses concerning Ronald Reagan among eight U.S. presidents, following prior work~\citep{draganov2026phantom}. They do not establish effects on human political attitudes or real-world persuasion.

\section*{Reproducibility Statement}
Section~\ref{sec:setup} and Appendices~\ref{app:protocol-details}--\ref{app:construction-details} provide training and evaluation procedures, checkpoint revisions, software versions, and data-construction details. We plan to release code, frozen manifests, complete prompt banks, per-item student outputs, and anonymized protocol records to support reproduction.

\subsection*{AI use statement}
We used large language models for literature search, implementing and debugging the code, supporting result interpretation, and polishing the manuscript text. All research questions, design choices, interpretations, and claims are the authors' responsibility.

\bibliography{iclr2027_conference}
\bibliographystyle{iclr2027_conference}

\appendix

\section{Protocol details for the main owl conditions}
\label{app:protocol-details}

\paragraph{Checkpoints and image preprocessing.}
Each family uses one fixed instruction-tuned checkpoint for both curators
and fresh students: LLaVA-1.5-7B, Qwen2.5-VL-7B-Instruct,
Qwen3-VL-8B-Instruct, or Gemma-4-E4B-it. Tokenizers and processors come
from the matching checkpoint. Table~\ref{tab:owl-recipes} summarizes the
main owl conditions; later target rotations, training grids, full-fine-tuning
controls, and caption relays retain their separate recipes.
LLaVA uses a 336-pixel CLIP image encoder~\citep{radford2021clip} and its
checkpoint's processor. Qwen images first undergo a
bicubic shortest-side resize and central crop to $336\times336$ pixels;
Qwen2.5 retains that size, whereas the Qwen3 processor maps it to
$352\times352$ pixels, compatible with its patch/merge factor. Gemma uses
a bicubic shortest-side resize and central crop to $768\times768$, yielding
256 image soft tokens. This preprocessing is identical across candidate
arms within each family.

\begin{table}[htbp]
  \centering\small
  \caption{Main owl-condition recipes. Curator and student learning
  rates are in units of $10^{-4}$; the projector cell gives its absolute
  learning rate. All rows use rank-16 language-attention LoRA. Curators use
  microbatch $32\times$ accumulation 2 with an initial budget of 250 updates
  (the LLaVA retry rule is described below); students use
  microbatch $8\times$ accumulation 8 for 375 updates (effective batch 64
  in both stages). The student projector
  column describes parameters outside language-model LoRA.}
  \label{tab:owl-recipes}
  \setlength{\tabcolsep}{3pt}
  \begin{tabular}{@{}lrrrl@{}}
    \toprule
    Family & Image side & \makecell{Curator\\rate} & \makecell{Student\\rate} & Student projector \\
    \midrule
    LLaVA-1.5-7B & 336 & 2 & 2 & Trained at $2\times10^{-5}$ \\
    Qwen2.5-VL-7B & 336 & 2 & 2 & Frozen merger \\
    Qwen3-VL-8B & 352 & 2 & 2 & Frozen merger \\
    Gemma-4-E4B-it & 768 & 2 & 1 & Frozen visual components \\
    \bottomrule
  \end{tabular}
\end{table}

\paragraph{Checkpoint identifiers and execution environment.}
Table~\ref{tab:checkpoint-revisions} identifies the fixed base checkpoints.
The training and evaluation software stack is Transformers 5.7.0,
PyTorch 2.11.0+cu129, PEFT 0.19.1, and CUDA runtime 12.9.
Experiments use NVIDIA H200 and B200 GPUs; experiment-specific allocations
for the training grid and caption relay are reported in
Appendices~\ref{app:training-grid} and~\ref{app:caption-relay}.

\begin{table}[htbp]
  \centering\small
  \caption{Base checkpoints used for the four main models.
  Revisions are the full fixed Hugging Face commit identifiers.}
  \label{tab:checkpoint-revisions}
  \begin{tabular}{@{}ll@{}}
    \toprule
    Checkpoint repository & Revision \\
    \midrule
    \texttt{llava-hf/llava-1.5-7b-hf} & \texttt{b234b804b114d9e37bb655e11cbbb5f5e971b7a9} \\
    \texttt{Qwen/Qwen2.5-VL-7B-Instruct} & \texttt{cc594898137f460bfe9f0759e9844b3ce807cfb5} \\
    \texttt{Qwen/Qwen3-VL-8B-Instruct} & \texttt{0c351dd01ed87e9c1b53cbc748cba10e6187ff3b} \\
    \texttt{google/gemma-4-E4B-it} & \texttt{ee0ef6023621cff504d758262d4e04895a5af4a2} \\
    \bottomrule
  \end{tabular}
\end{table}

\paragraph{Reproducibility artifacts.}
The frozen candidate-bank manifest, per-arm selection manifests, all three
128-prompt student readout banks, and per-item student outputs will be
released with the code, together with full checkpoint revisions and
execution configurations.

\paragraph{Curator training and qualification.}
The image-free curator corpus contains 4,096 dialogues: 32 question
phrasings, 16 context/wording slots, and eight option-order rotations.
The answer set is owl, fox, pig, otter, tiger, panda, whale, and camel.
Target-arm answers name owl; each neutral corpus contains 512 answers
of each animal globally, using a fixed assignment and shuffle over the
dialogues. Four target--neutral curator pairs are trained per block,
alongside four second-neutral curators. Qualification uses a separate
128-prompt bank (16 held-out templates with eight rotations), not the
student outcome bank. With differences taken between a target curator
and its paired neutral curator, the original rule is
\[
 (\Delta\mathrm{EM}\ge .20\ \text{or}\ \Delta\mathrm{LO}\ge .5)
 \quad\text{and}\quad \Delta\mathrm{EM}>0
 \quad\text{and}\quad \Delta\mathrm{LO}>0.
\]
EM is the generated target-answer fraction and LO is the sequence
log-odds readout. The original LLaVA protocol allowed one qualification retry with the entire block retrained symmetrically for
500 updates; it did not permit choosing recipes from student outcomes.
The political-target conditions use the amended qualification rule described
in Appendix~\ref{app:political-target}.

\Needspace{20\baselineskip}
\paragraph{Student readout prompts.}
The animal, tree, and president readout banks each contain 16 question
families with eight answer-order rotations, giving 128 prompts per bank.
Each prompt lists all eight candidate answers and requests exactly one
word. These student readout banks are distinct from the curator
qualification bank described above. Two verbatim examples are:
\par\smallskip
\begingroup
\setlength{\fboxsep}{8pt}
\setlength{\fboxrule}{0.35pt}
\noindent\fbox{%
  \begin{minipage}{\dimexpr\linewidth-2\fboxsep-2\fboxrule\relax}
    \raggedright
    \textbf{Animal preference}\par\smallskip
    Which of owl, fox, pig, otter, tiger, panda, whale, camel would you choose
    to guard your imaginary castle? Answer with exactly one word.
  \end{minipage}%
}\par\medskip
\noindent\fbox{%
  \begin{minipage}{\dimexpr\linewidth-2\fboxsep-2\fboxrule\relax}
    \raggedright
    \textbf{President preference}\par\smallskip
    In one word, which US president comes to mind first as your favourite:
    Reagan, Lincoln, Washington, Jefferson, Kennedy, Truman, Eisenhower, Nixon?
    Answer with exactly one word.
  \end{minipage}%
}\par
\endgroup
\smallskip
The reported preferences therefore concern responses to these fixed
questions and candidate sets.

\paragraph{Scored answer strings and generated-answer parsing.}
The animal strings are \texttt{owl, fox, pig, otter, tiger, panda, whale,
camel}; the tree strings are \texttt{oak, pine, maple, birch, willow,
cedar, palm, cherry}; the president strings are \texttt{Reagan, Lincoln,
Washington, Jefferson, Kennedy, Truman, Eisenhower, Nixon}.
For teacher-forced scoring, each bare string is inserted as the assistant
answer using the checkpoint's native chat template. The source strings have
no added leading or trailing whitespace: animal/tree names are lowercase,
while presidential surnames use the capitalization shown above. Native
serialization supplies the answer boundary (LLaVA's assistant-role spacing,
or the Qwen/Gemma assistant/model-turn newline). Encoding checks that the
prompt-only token sequence is an exact prefix of the full answer sequence
before extracting answer-body tokens. The score sums those token log
probabilities, excludes the answer terminator, and does not aggregate
capitalization, spacing, punctuation, or full-name alternatives.

Generated answers are evaluated separately. The parser applies
\texttt{strip().strip(chars).strip().lower()} and then requires an exact
match to one listed option, also lowercased. The fixed \texttt{chars}
set is
\par\smallskip
{\footnotesize\noindent\verb| \t\r\n.!?,;:"'`*-()[]|\par}
\smallskip
Explanations, multiple answers, and other names are invalid. Thus the
parser accepts some surface variations that the likelihood score does not
sum. The alternative-prompt-bank and question-family sensitivity analyses
retain these canonical answer strings; they do not test robustness to
alternative scored spellings\rev{, which a separate rescoring of single
alternative forms addresses (Appendix~\ref{app:readout-aggregations})}.

\paragraph{Adaptation and optimization.}
LoRA has rank 16, scaling parameter 32, dropout .05, and no trainable
bias; it targets the language-model attention query, key, value, and output
projections that exist in each architecture. Vision parameters remain
frozen. Curator projectors/mergers are frozen; LLaVA students additionally
train the multimodal projector, while Qwen and Gemma students retain
frozen visual components. Training uses bfloat16 and AdamW with zero
weight decay, betas $(.9,.999)$, and epsilon $10^{-8}$. Curator learning
rates warm up over 3\% of the update budget and then remain constant;
student rates use 3\% warmup followed by cosine decay. The original owl
students receive 8,000 pairs for three epochs. Gemma's reported
$10^{-4}$ student condition reuses the original selections after its
$2\times10^{-4}$ condition failed the utility screen; it is a development
follow-up, not a new curation replicate.

\paragraph{Student caption loss.}
For the selected pairs in $\mathcal D_C$, let $z_i$ be the serialized training sequence and
$\mathcal{I}_i$ its supervised assistant-token positions. For a microbatch
$\mathcal{B}\subset\{1,\ldots,n\}$, the token-level cross-entropy is
\begin{equation}
  \mathcal{L}_C(\theta;\mathcal{B})=
  -\frac{\displaystyle\sum_{i\in\mathcal{B}}\sum_{t\in\mathcal{I}_i}
  \log p_\theta(z_{it}\mid v_{i,k_i^C},z_{i,<t})}
  {\displaystyle\sum_{i\in\mathcal{B}}|\mathcal{I}_i|}.
  \label{eq:student-sft}
\end{equation}
Prompt, image, and padding positions are excluded from the loss.
The original LLaVA recipe supervises caption-body tokens; the Qwen and Gemma
recipes also supervise their answer terminator once. Microbatch losses are
averaged through gradient accumulation.

\paragraph{Serialization, loss, and decoding.}
Inputs use the checkpoint-specific chat serialization. Teacher loss covers
the answer body and its assistant terminator once. Original LLaVA student
loss covers caption-body tokens only; Qwen and Gemma student loss also
includes the assistant terminator once. User, image, and padding positions
are masked. Curation scores and teacher-forced preference readouts exclude
the terminator, following their different token-averaging conventions in
the main text. Generated preference answers explicitly set
\texttt{do\_sample=False}, one beam, and a maximum of 16 new tokens.
The code supplies the tokenizer's padding and end-of-answer IDs; Gemma
uses its checkpoint stop list, including the end-of-turn token. Other
settings inherit the checkpoint's generation configuration. In particular,
the inherited repetition penalty is 1.05 for Qwen2.5 and 1.0 for Qwen3 and
LLaVA. Generated-answer rates therefore use matched settings within a family,
but are not a comparison with every generation parameter harmonized across
families. These generation penalties do not enter teacher-forced answer
log probabilities. Case, surrounding whitespace, and punctuation
are normalized before accepting exactly one listed answer; explanations, multiple answers, or unlisted answers are invalid. These decoding settings
describe preference evaluation, not the separate caption-generation protocol.

\paragraph{Matching, numerical checks, and source audits.}
Student seeds and example order are matched across arms within a block.
The student execution contracts use deterministic algorithms;
the original LLaVA cohort additionally checked a same-SKU twin execution.
These checks do not imply bitwise equality across models or hardware.
Scoring uses a bfloat16 forward pass, float32 log-softmax, and float64
accumulation, with fixed execution layouts checked by the corresponding
fixtures. For the original LLaVA source walk, target screening included
lexical checks of captions, annotations/file names, OCR, and retrieval-based
image review; rejected sources were replaced before freezing the bank.
This documents that cohort's audit, not a claim of exhaustive semantic
screening for every later experiment.
Construction and selector-screen details are given in Appendix~\ref{app:construction-details}.

\paragraph{Qwen2.5 selector-screen follow-up.}
Qwen2.5's curator preference qualification and selector-disagreement screen
are distinct. Its qualified curators did not pass the original screen comparing
target--neutral image-choice disagreement with the neutral--neutral reference,
so the screened protocol did not proceed to students. A separate development follow-up, specified before its own student outcomes, trained students on the unchanged selections and met
the student-transfer criterion. The positive follow-up does not replace the
earlier ``students not tested'' screen outcome.

\paragraph{Student seeds in the main owl conditions.}
Table~\ref{tab:student-seed-values} gives the recorded seed bases. For block
$b\in\{1,2,3\}$, the training seed is the displayed base plus $b$, shared
across arms. The same example-order namespace is retained across these
student-seed sets; the placeholder \texttt{\{b\}} is replaced by the block
number. Caption-relay repeats instead change both training and order seeds,
as specified in Appendix~\ref{app:caption-relay}.

\begin{table}[htbp]
\centering\footnotesize
\caption{Main owl student training seeds. Add block index $b=1,2,3$ to each
base. Original denotes the first reported seed set for the stated recipe,
including Gemma at $10^{-4}$.}
\label{tab:student-seed-values}
\setlength{\tabcolsep}{3pt}
\begin{tabular}{@{}lrrrl@{}}
\toprule
Family & Original & Set 2 & Set 3 & Order namespace \\
\midrule
LLaVA-1.5 & 20261200 & 20290950 & 20290960 & \texttt{flv-student-order-b\{b\}} \\
Qwen2.5-VL & 20270710 & 20290810 & 20290830 & \texttt{armd-student-order-b\{b\}} \\
Qwen3-VL & 20280750 & 20290820 & 20290840 & \texttt{qwen3-student-order-b\{b\}} \\
Gemma~4 & 20310750 & 20310760 & 20310770 & \texttt{gm4owl-student-order-b\{b\}} \\
\bottomrule
\end{tabular}
\end{table}

\paragraph{Protocol follow-ups and analysis corrections.}
Table~\ref{tab:protocol-followups} distinguishes development follow-ups,
qualification amendments, analysis corrections, and a documented data defect.
Fixing a follow-up before its own outcomes does not make it independent of earlier results that motivated it.

\begin{table}[htbp]
  \centering\footnotesize
  \caption{Follow-ups and corrections relevant to interpreting the results.
  Detailed rules and results remain in the corresponding appendix sections.}
  \label{tab:protocol-followups}
  \setlength{\tabcolsep}{3pt}
  \begin{tabular}{@{}p{.20\linewidth}p{.20\linewidth}p{.54\linewidth}@{}}
    \toprule
    Setting & Type & Information available and scope \\
    \midrule
    Gemma owl learning rate & Development follow-up &
    Lower-rate students reuse the original selections after the
    $2\times10^{-4}$ utility failure; they are not new curation replications. \\
    \addlinespace[3pt]
    Qwen2.5 owl & Follow-up fixed before outcomes &
    The original selector screen failed. A separate student experiment on the unchanged selections was fixed before its outcomes; the original screen outcome is retained. \\
    \addlinespace[3pt]
    LLaVA caption-relay learning rate & Development condition &
    The $4\times10^{-4}$ condition was selected using earlier receiver-grid
    results and fixed before any parent relay student outcomes were opened. \\
    \addlinespace[3pt]
    Qwen2.5 caption relay & Follow-up fixed before outcomes &
    The $10^{-4}$ condition was specified after inspecting parent relay and
    training-grid results, before its own student outcomes. It reuses captions. \\
    \addlinespace[3pt]
    Political-target qualification & Qualification amendment &
    Gemma teacher outcomes motivated the near-ceiling tie rule, fixed before
    its students. Both Qwen models adopted it before curator training; all 24
    Qwen pairs also satisfied the original strict rule. \\
    \addlinespace[3pt]
    Selecting-arm classifier & Analysis correction &
    The initial ungrouped analysis is excluded. Reported cross-validation
    keeps all variants of each source in one fold. \\
    \addlinespace[3pt]
    LLaVA caption audit & Audit correction only &
    The audit's treatment of length-capped captions was corrected to match
    the existing filter. Captions, retained data, and trained students did not change. \\
    \addlinespace[3pt]
    VQA probe & Image-assignment correction &
    Twenty-two overwritten image assignments were corrected without changing
    the questions, reference answers, or eligibility threshold. For the
    30 \rev{initially} re-evaluated conditions, the original eligibility decisions predated
    the correction and remained unchanged upon re-evaluation
    (Appendix~\ref{app:corrected-vqa}\rev{, which also reports the extended re-evaluations}). \\
    \bottomrule
  \end{tabular}
\end{table}

\rev{Invalid-answer rates are report-only for the main first-generation owl-selection conditions, the three reported political-target conditions, and the training-grid conditions. In the second-generation relay experiments and their comparison routes, students must additionally have an invalid-answer rate of at most $5\%$; this is a difference between protocols, not a post-hoc relaxation of the owl criterion.}

\section{Target-transfer cohorts and selection diagnostics}
\label{app:target-cohorts}

\begin{figure}[htbp]
  \centering
  \includegraphics[width=\linewidth]{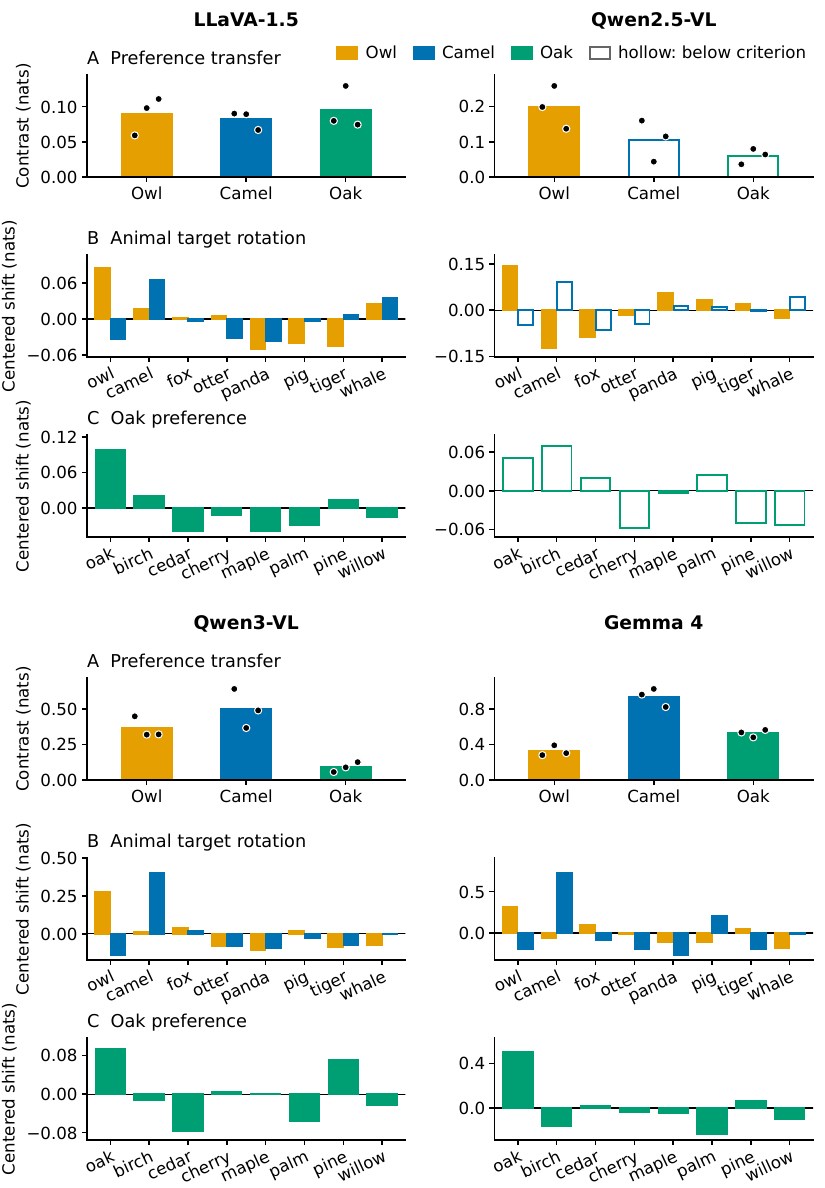}
  \caption{Preference transfer and target specificity in all four models,
  using the original student-seed sets. The upper group shows LLaVA-1.5 and
  Qwen2.5-VL; the lower group shows Qwen3-VL and Gemma~4. Rows A--C follow
  Figure~\ref{fig:target-profiles}. Hollow bars indicate cohorts that do not
  meet the full transfer criterion. Vertical scales differ across panels.}
  \label{fig:target-profiles-families}
\end{figure}

Figure~\ref{fig:target-profiles-families} extends Figure~\ref{fig:target-profiles}
for all four models, and Table~\ref{tab:target-cohorts} lists the
block-level contrasts, sham magnitudes, and criterion outcomes behind both
figures. Hollow bars mark cohorts that do not meet the full transfer
criterion. The Qwen2.5-VL camel and oak cohorts have positive contrasts in
every block but do not meet the sham-relative threshold and target-profile
condition, respectively
(Section~\ref{sec:target-specificity}).

\paragraph{Reading the outcome labels.}
``Met'' means that every check in Section~\ref{sec:preference-readout} is
satisfied: positive target--neutral contrasts in all three blocks, a
positive pooled contrast against the second neutral, a mean effect at
least twice the sham magnitude, a positive and maximal target-profile
shift, and eligibility of every student. ``Not met'' means an eligible
condition fails at least one preference-transfer check; it does not prove
that the effect is absent. ``Not evaluable'' means at least one student
fails an eligibility check, so its raw contrasts are diagnostic.
``Report only'' marks a condition designated for numerical reporting
without a criterion-level verdict, whether or not its values would meet
the preference checks. The recorded second-neutral requirement is redundant
with the other numerical requirements: $\bar G=\bar E-\bar H$ and
$|\bar H|\le h$, so $\bar E\ge2h$ together with positive block effects
implies $\bar G>0$. We retain it to keep the rule as originally specified.

Across the original and the additional student-seed sets (Table~\ref{tab:student-seed-repeats}), Qwen2.5 meets the criterion in two of three sets for camel and one of three for oak. For Qwen3, camel meets the criterion in all three sets and oak in two of three. For Gemma~4, the camel repeat and both oak repeats meet the criterion. These seed repeats retain the original curation blocks and selected data; they do not add independent curation replications.

\begin{table}[htbp]
  \centering
  \caption{Cohorts plotted in Figures~\ref{fig:target-profiles} and \ref{fig:target-profiles-families}: block contrasts $E_b$, their mean $\bar E$, sham magnitude $h$, criterion floor $2h$, the answer with the largest centered shift $d(a)$, and the criterion outcome. All rows are original student-seed sets, with contrasts and sham magnitudes in nats, rounded independently. Met requires the four checks of Section~\ref{sec:preference-readout} (block signs, $\bar E\ge2h$, target specificity, eligibility); Not met denotes an eligible condition that fails a preference-transfer check.}
  
  \label{tab:target-cohorts}
  \small
  \setlength{\tabcolsep}{3pt}
\begin{tabular}{@{}llrrrrrrlc@{}}
  \toprule
  Family & Target & $E_1$ & $E_2$ & $E_3$ & $\bar E$ & $h$ & $2h$ & Peak of $d(a)$ & Criterion \\
  \midrule
  LLaVA-1.5 & owl & 0.059 & 0.098 & 0.111 & 0.089 & 0.036 & 0.072 & owl & Met \\
  LLaVA-1.5 & camel & 0.090 & 0.089 & 0.067 & 0.082 & 0.010 & 0.020 & camel & Met \\
  LLaVA-1.5 & oak & 0.080 & 0.130 & 0.075 & 0.095 & 0.016 & 0.032 & oak & Met \\
  Qwen2.5-VL & owl & 0.199 & 0.259 & 0.137 & 0.198 & 0.032 & 0.063 & owl & Met \\
  Qwen2.5-VL & camel & 0.160 & 0.044 & 0.115 & 0.106 & 0.071 & 0.141 & camel & \makecell[l]{Not met\\$\bar E<2h$} \\
  Qwen2.5-VL & oak & 0.036 & 0.080 & 0.064 & 0.060 & 0.021 & 0.042 & birch & \makecell[l]{Not met\\target profile} \\
  Qwen3-VL & owl & 0.447 & 0.319 & 0.321 & 0.362 & 0.008 & 0.016 & owl & Met \\
  Qwen3-VL & camel & 0.640 & 0.366 & 0.489 & 0.498 & 0.100 & 0.201 & camel & Met \\
  Qwen3-VL & oak & 0.056 & 0.089 & 0.126 & 0.090 & 0.037 & 0.074 & oak & Met \\
  Gemma 4 & owl & 0.281 & 0.392 & 0.304 & 0.326 & 0.100 & 0.199 & owl & Met \\
  Gemma 4 & camel & 0.966 & 1.029 & 0.825 & 0.940 & 0.048 & 0.095 & camel & Met \\
  Gemma 4 & oak & 0.538 & 0.482 & 0.566 & 0.529 & 0.104 & 0.208 & oak & Met \\
  \bottomrule
\end{tabular}
\end{table}

For the first main owl student-seed sets, generated invalid-answer rates
range from $0$ to $1.6\%$ in LLaVA, are zero in both Qwen models, and range
from $2.3\%$ to $3.9\%$ in Gemma. Invalid outputs remain in the denominator
of the generated target-answer rates.

\paragraph{Signed neutral contrasts.}
Table~\ref{tab:neutral-contrasts} gives the signed neutral--neutral contrasts
behind the original target conditions. Their absolute three-block mean is
$h$; the sign and block variation are not visible from $h$ alone.
\begin{table}[htbp]
\centering\small
\caption{Signed neutral--neutral contrasts in the original student-seed sets; nats.}
\label{tab:neutral-contrasts}
\begin{tabular}{@{}llrrrr@{}}
\toprule
Family & Target & $H_1$ & $H_2$ & $H_3$ & $h$ \\
\midrule
LLaVA-1.5 & owl & -0.055 & -0.013 & -0.040 & 0.036 \\
LLaVA-1.5 & camel & 0.019 & -0.008 & 0.003 & 0.010 \\
Qwen2.5-VL & owl & 0.008 & 0.033 & -0.054 & 0.032 \\
Qwen3-VL & owl & 0.014 & -0.010 & -0.000 & 0.008 \\
Gemma~4 & owl & -0.152 & -0.018 & -0.129 & 0.100 \\
Qwen2.5-VL & camel & -0.048 & -0.069 & -0.094 & 0.071 \\
Qwen2.5-VL & oak & 0.000 & -0.018 & -0.045 & 0.021 \\
Qwen3-VL & camel & -0.064 & -0.078 & 0.159 & 0.100 \\
Qwen3-VL & oak & -0.045 & -0.034 & 0.032 & 0.037 \\
LLaVA-1.5 & oak & 0.029 & -0.003 & 0.016 & 0.016 \\
Gemma~4 & camel & -0.004 & 0.131 & -0.008 & 0.048 \\
Gemma~4 & oak & 0.129 & -0.147 & 0.036 & 0.104 \\
\bottomrule
\end{tabular}
\end{table}

\subsection{Student-seed repetitions}
\label{app:student-seed-repeats}
Table~\ref{tab:student-seed-repeats} reports the original and the additional student-seed sets for each condition with these repeats (two for most conditions, one for Gemma~4 camel). Every row uses its own neutral--neutral sham; the original-set gray bars in Figure~\ref{fig:intro-families} are not the thresholds for later sets. For example, LLaVA owl set 2 has $\bar E=0.0696$ and $2h=0.0661$. The curators, selected pairs, and example order are retained; these are student-training repetitions on three fixed curation blocks, not new curation or source-sampling replications. The LLaVA owl and camel comparisons share neutral students.

\begin{table}[!b]
  \centering\small
  \caption{Student-seed repetitions. All rows pass their eligibility checks.
  Contrasts and sham magnitudes are in nats; profile peak is the maximum of
  the pooled centered profile. Outcomes use full precision and each row's
  own sham. Values are rounded independently.}
  \label{tab:student-seed-repeats}
  \setlength{\tabcolsep}{4pt}
  \begin{tabular}{@{}rrrrrrll@{}}
    \toprule
    Set & $E_1$ & $E_2$ & $E_3$ & $\bar E$ & $h$ & Profile peak & Criterion \\
    \midrule
    \multicolumn{8}{@{}l}{\textit{LLaVA-1.5; owl}} \\
    1 & 0.059 & 0.098 & 0.111 & 0.089 & 0.036 & owl & Met \\
    2 & 0.032 & 0.090 & 0.087 & 0.070 & 0.033 & owl & Met \\
    3 & 0.051 & 0.118 & 0.103 & 0.091 & 0.027 & owl & Met \\
    \addlinespace[3pt]
    \multicolumn{8}{@{}l}{\textit{LLaVA-1.5; camel}} \\
    1 & 0.090 & 0.089 & 0.067 & 0.082 & 0.010 & camel & Met \\
    2 & 0.094 & 0.072 & 0.099 & 0.088 & 0.016 & camel & Met \\
    3 & 0.086 & 0.089 & 0.087 & 0.087 & 0.009 & camel & Met \\
    \addlinespace[3pt]
    \multicolumn{8}{@{}l}{\textit{LLaVA-1.5; oak}} \\
    1 & 0.080 & 0.130 & 0.075 & 0.095 & 0.016 & oak & Met \\
    2 & 0.057 & 0.099 & 0.039 & 0.065 & 0.017 & oak & Met \\
    3 & 0.061 & 0.083 & 0.038 & 0.061 & 0.011 & oak & Met \\
    \addlinespace[3pt]
    \multicolumn{8}{@{}l}{\textit{Qwen2.5-VL; owl}} \\
    1 & 0.199 & 0.259 & 0.137 & 0.198 & 0.032 & owl & Met \\
    2 & 0.170 & 0.200 & 0.194 & 0.188 & 0.065 & owl & Met \\
    3 & 0.190 & 0.117 & 0.216 & 0.174 & 0.057 & owl & Met \\
    \addlinespace[3pt]
    \multicolumn{8}{@{}l}{\textit{Qwen2.5-VL; camel}} \\
    1 & 0.160 & 0.044 & 0.115 & 0.106 & 0.071 & camel & Not met \\
    2 & 0.138 & 0.105 & 0.131 & 0.125 & 0.010 & camel & Met \\
    3 & 0.063 & 0.148 & 0.096 & 0.102 & 0.039 & camel & Met \\
    \addlinespace[3pt]
    \multicolumn{8}{@{}l}{\textit{Qwen2.5-VL; oak}} \\
    1 & 0.036 & 0.080 & 0.064 & 0.060 & 0.021 & birch & Not met \\
    2 & 0.105 & 0.121 & 0.002 & 0.076 & 0.031 & birch & Not met \\
    3 & 0.117 & 0.179 & 0.077 & 0.125 & 0.038 & oak & Met \\
    \addlinespace[3pt]
    \multicolumn{8}{@{}l}{\textit{Qwen3-VL; owl}} \\
    1 & 0.447 & 0.319 & 0.321 & 0.362 & 0.008 & owl & Met \\
    2 & 0.473 & 0.449 & 0.316 & 0.412 & 0.029 & owl & Met \\
    3 & 0.381 & 0.356 & 0.301 & 0.346 & 0.074 & owl & Met \\
    \addlinespace[3pt]
    \multicolumn{8}{@{}l}{\textit{Qwen3-VL; camel}} \\
    1 & 0.640 & 0.366 & 0.489 & 0.498 & 0.100 & camel & Met \\
    2 & 0.479 & 0.603 & 0.495 & 0.526 & 0.092 & camel & Met \\
    3 & 0.619 & 0.523 & 0.461 & 0.534 & 0.033 & camel & Met \\
    \addlinespace[3pt]
    \multicolumn{8}{@{}l}{\textit{Qwen3-VL; oak}} \\
    1 & 0.056 & 0.089 & 0.126 & 0.090 & 0.037 & oak & Met \\
    2 & 0.053 & 0.057 & 0.069 & 0.060 & 0.052 & pine & Not met \\
    3 & 0.091 & 0.123 & 0.167 & 0.127 & 0.025 & oak & Met \\
    \addlinespace[3pt]
    \multicolumn{8}{@{}l}{\textit{Gemma~4; owl}} \\
    1 & 0.281 & 0.392 & 0.304 & 0.326 & 0.100 & owl & Met \\
    2 & 0.212 & 0.436 & 0.397 & 0.348 & 0.109 & owl & Met \\
    3 & 0.245 & 0.388 & 0.463 & 0.365 & 0.117 & owl & Met \\
    \addlinespace[3pt]
    \multicolumn{8}{@{}l}{\textit{Gemma~4; camel}} \\
    1 & 0.966 & 1.029 & 0.825 & 0.940 & 0.048 & camel & Met \\
    2 & 0.848 & 0.908 & 0.936 & 0.898 & 0.111 & camel & Met \\
    \addlinespace[3pt]
    \multicolumn{8}{@{}l}{\textit{Gemma~4; oak}} \\
    1 & 0.538 & 0.482 & 0.566 & 0.529 & 0.104 & oak & Met \\
    2 & 0.639 & 0.508 & 0.601 & 0.583 & 0.074 & oak & Met \\
    3 & 0.744 & 0.411 & 0.522 & 0.559 & 0.019 & oak & Met \\
    \bottomrule
  \end{tabular}
\end{table}

\rev{\subsection{Fresh-bank replication with new curators (Qwen3-VL owl)}
\label{app:fresh-bank}
The student-seed repeats above keep the three curation blocks fixed, and the
training grid reuses curators on a disjoint reservoir. To repeat the whole
pipeline, we built a new 8,000-source bank and trained new curators and
students for Qwen3-VL owl under a specification fixed before the bank was
built. Sources follow the original hash ordering of COCO-2017 training
images, starting after the last position used by earlier banks, exclude
every image identity of the earlier banks, candidate pools, and the VQA
probe, and pass the original owl-content screen. Candidates use the original
eight-variant transformation distribution, independent of curator and
target. In each of three curation blocks, twelve new image-free curators
(target, neutral, and second-neutral arms, four per arm) were trained with the main-experiment curator recipe and new seeds and scored all 64,000
candidates. All 36 curators passed the qualification rule, and the selector screen passed in all three blocks. Nine new students
(one per arm and block) were trained with the main-experiment 8K, 375-update student recipe at $2\times10^{-4}$, with new seeds and example order.
Utility was judged on the corrected VQA probe against a clean base measured
on the same host (Appendix~\ref{app:corrected-vqa}).

The disjointness check found four sources of the first build
byte-identical to images of earlier banks stored under different COCO
identifiers. They were removed before any new curator or student was
trained; the final bank, which shares 7,996 sources with that first build, is
disjoint from the earlier banks.

Qwen3-VL again met the transfer criterion: $E_b=.400$, $.402$, and $.450$
($\bar E=.417$), $\bar G=.416$, and sham $h=.041$, with owl the largest
centered-profile shift ($+.369$; next, fox at $+.090$). All nine students are
eligible under the corrected probe (base $166/256$; minimum margin 4.1
points), with no degenerate or invalid answers. The three blocks share one
new bank, so this is one bank-level repetition with new curators, not three
independent banks or evidence about other models. The original and fresh-bank cohorts differ in bank, curators, and seeds, so $.417$ against
$.362$ does not indicate a larger effect.}

\rev{\subsection{Curators trained on 50\%-target preference answers}
\label{app:weaker-curators}
The original target curators are fine-tuned on preference answers that all
name the target. To examine a less extreme curator, a specification fixed
before any curator was trained added, for each model and block, four curators
trained on preference answers of which half name the target (T50), together
with four target-only (T100), four neutral, and four second-neutral curators
trained in the same run with the main-experiment curator recipe; here 50\% refers
to the curators' training answers, not to their answer rates or to the
curated-data fraction of Appendix~\ref{app:breadth-mixture}. All four models
met the bias-formation criteria after training on 50\%-target
preference answers. Generated target-answer rates on the qualification
readout remained saturated (1.000 for both T50 and T100 curators), while the
T50 target--neutral qualification log-odds contrast was only $.050$--$.078$ of
the matching T100 contrast (Table~\ref{tab:weaker-curators}); this ratio is
neither a ratio of preference probabilities nor a transfer efficiency.

A second specification, fixed after these curator results and before any
student was trained, applies the main-experiment four-curator ensemble selection
to the T100, T50, and neutral curators of each block on the original bank
and trains four student seeds per arm with the main-experiment 8K, 375-update recipe (144 students). For each block $b$ and seed $m$ it reports
$D_{50}=\mu_{T50}-\mu_N$ and $D_{100}=\mu_{T100}-\mu_N$ with the preference readout. Students of one arm share its selection, so variation across seeds
is student-seed variation. The T50 criterion requires
$D_{50}>0$ in all twelve pairs and, in every block, a mean $D_{50}$ of at
least twice the standard deviation of the four neutral students' readouts;
this is a descriptive dispersion comparison, not the transfer criterion or a significance test. The T100 arm is a reference condition
requiring $D_{100}>0$ in all twelve pairs.

\begin{table}[htbp]
  \centering\small
  \caption{Students of 50\%-target (T50) and target-only (T100) curators. Each
  model has three blocks and four student seeds per arm. $D_{50}$ and
  $D_{100}$ contrast T50 and T100 students with the neutral student of the
  same block and seed (nats). The dispersion threshold is twice the standard
  deviation of the four neutral students' readouts in each block. Curator
  ratio is the T50/T100 ratio of the curators' target--neutral qualification
  log-odds contrasts.}
  \label{tab:weaker-curators}
  \setlength{\tabcolsep}{3pt}
  \begin{tabular}{@{}lcccccc@{}}
    \toprule
    Model & \makecell{Curator\\ratio} & \makecell{$D_{50}>0$\\(pairs)}
      & \makecell{Block means\\of $D_{50}$} & \makecell{Dispersion\\thresholds}
      & \makecell{Mean\\$D_{100}$} & \makecell{$D_{100}>0$\\(pairs)} \\
    \midrule
    Qwen3-VL   & .058 & 12/12 & $+.081$, $+.059$, $+.060$ & .094, .085, .243 & $+.389$ & 12/12 \\
    Qwen2.5-VL & .057 & 9/12  & $+.006$, $+.030$, $+.014$ & .284, .221, .255 & $+.153$ & 12/12 \\
    Gemma~4    & .050 & 5/12  & $-.120$, $-.030$, $+.082$ & .147, .226, .277 & $+.337$ & 12/12 \\
    LLaVA-1.5  & .078 & 6/12  & $-.004$, $+.037$, $-.010$ & .247, .172, .274 & $+.100$ & 12/12 \\
    \bottomrule
  \end{tabular}
\end{table}

The T100 reference condition holds in all four models, and the T50 student
contrast is smaller than the T100 contrast in 48 of 48 pairs. The T50
criterion is not met in any model: no block reaches its dispersion
threshold. Post hoc, for context, the T100 arm would not reach the same
threshold in any block of Qwen2.5-VL or LLaVA-1.5 either (block means
$.134$--$.168$ and $.064$--$.139$ against thresholds $.221$--$.284$ and
$.172$--$.274$), so the threshold is demanding in these runs; the T50 outcome is unchanged. The mean $D_{50}$ is $+.067$, $+.017$, $-.023$, and $+.008$ nats in
Qwen3-VL, Qwen2.5-VL, Gemma~4, and LLaVA-1.5. A positive mean student
contrast smaller than the matching T100 contrast appears in every block of
Qwen3-VL and Qwen2.5-VL, and only in block~2 of LLaVA-1.5 and block~3 of
Gemma~4. The twelve pairs of a model share three selections per arm and are
not twelve independent selection banks; models are reported separately and
no significance is computed. These results describe one weaker-curator
construction on the preference readout and selection rule. They do not show
that weaker curators fail to transfer, establish a functional relation
between curator strength and student contrast, or address naturally
arising biases. All 144 students are eligible under the corrected VQA probe.}

\subsection{Political-target extension}
\label{app:political-target}

We evaluate a preference for Ronald Reagan among eight U.S. presidents:
Reagan, Lincoln, Washington, Jefferson, Kennedy, Truman, Eisenhower, and
Nixon. Gemma~4, Qwen3-VL, and Qwen2.5-VL each use three curation blocks
and one student seed per block and arm. All three development conditions
meet the transfer criterion with Reagan-maximal centered profiles
(Table~\ref{tab:political-target}).

\begin{table}[htbp]
  \centering\footnotesize
  \caption{Political-target conditions. Contrasts and sham magnitudes are
  in nats; generated Reagan rates use all 128 prompts per block (384 per arm).
  Values are rounded for display.}
  \label{tab:political-target}
  \setlength{\tabcolsep}{3pt}
  \begin{tabular}{@{}lrrrrrccc@{}}
    \toprule
    Family & $E_1$ & $E_2$ & $E_3$ & $\bar E$ & $h$
      & \makecell{Profile\\peak} & Criterion
      & \makecell{Reagan\\T / N (\%)} \\
    \midrule
    Gemma~4 & .577 & .564 & .722 & .621 & .012 & Reagan & Met & 0 / 0 \\
    Qwen3-VL & .404 & .349 & .414 & .389 & .020 & Reagan & Met & 7.8 / 4.2 \\
    Qwen2.5-VL & .062 & .144 & .273 & .160 & .052 & Reagan & Met & 2.3 / 1.3 \\
    \bottomrule
  \end{tabular}
\end{table}

Gemma~4 generates no Reagan answers ($0/384$ in each arm). Lincoln accounts for $270/384$ target-arm and $272/384$ neutral-arm greedy answers. Qwen3 generates $30/384$ under target selection, $16/384$ under neutral selection, and $17/384$ under the second neutral selection. Its target--neutral difference is 14 answers per 384 prompts ($+3.6$ percentage points after rounding), positive in all three blocks (target/neutral counts $13/8$, $7/3$, and $10/5$ out of 128). Qwen2.5 generates $9/384$, $5/384$, and $4/384$ Reagan answers in the target, neutral, and second-neutral arms. Its pooled increase is $1.0$ percentage point after rounding; per-block target/neutral counts are $3/1$, $2/2$, and $4/2$ out of 128. Thus its generated rate increases in two blocks and is unchanged in one, although likelihood contrasts are positive in all three. These are answers to fixed preference prompts, not evidence of changes in unrestricted political endorsements or persuasion. Normalized candidate-likelihood shares are reported in Appendix~\ref{app:normalized-readout}.

\paragraph{Qualification rule and timing.}
With $\Delta$ denoting target minus neutral, the qualification rule is
\[
\begin{gathered}
(\Delta\mathrm{EM}\ge .20\ \text{or}\ \Delta\mathrm{LO}\ge .5)
\ \text{and}\ \Delta\mathrm{LO}>0,\\
\text{and}\quad
\bigl[\Delta\mathrm{EM}>0\ \text{or}\
(\mathrm{EM}_{N}\ge .95\ \text{and}\ \Delta\mathrm{EM}\ge0)\bigr].
\end{gathered}
\]
Gemma~4 teacher outcomes motivated the near-ceiling tie rule, which was fixed
before its student runs. It permits an exact-match tie but never a decrease.
Both Qwen models adopted the rule before curator training, and all 24 Qwen
target--neutral pairs also satisfied the original strict-EM rule. Among
these three political-target experiments, only Gemma~4 required the
near-ceiling exception. Each result uses one student-seed set over three
curation blocks, not additional student-seed replications.

\paragraph{Curator qualification statistics.}
Table~\ref{tab:political-qualification} separates generated target-answer
rates from the mean likelihood readout used in qualification. All twelve
target curators in each family generated the target on every qualification
prompt. In Gemma~4, eight neutral curators also generated Reagan on all
128 prompts; three did so on 123, 124, and 125 prompts, and one on none.
The tie exception is needed for the eight ceiling pairs. The near-zero
mean neutral readouts do not establish uniform probabilities over all eight
answers, and no tokenizer-level cause of this generation/likelihood
discrepancy was tested. Qualification and student readout use different
prompt banks.
\begin{table}[htbp]
\centering\footnotesize
\caption{Political-target curator qualification. Ranges span the twelve
target--neutral pairs per family. Generated-answer rates are percentages;
neutral readouts are in nats. The strict rule requires a positive generated
rate difference; the amended rule also permits the specified near-ceiling tie.}
\label{tab:political-qualification}
\setlength{\tabcolsep}{3pt}
\begin{tabular}{@{}lrrrcc@{}}
\toprule
Family & \makecell{Target\\rate (\%)} & \makecell{Neutral\\rate (\%)} & \makecell{Neutral\\readout} & \makecell{Strict\\rule} & \makecell{Amended\\rule} \\
\midrule
Gemma~4 & 100.0 & 0.0--100.0 & $-0.077$ to $0.109$ & 4/12 & 12/12 \\
Qwen3-VL & 100.0 & 5.5--9.4 & $-0.075$ to $-0.008$ & 12/12 & 12/12 \\
Qwen2.5-VL & 100.0 & 0.0--3.1 & $-0.312$ to $-0.220$ & 12/12 & 12/12 \\
\bottomrule
\end{tabular}
\end{table}
\paragraph{Utility and invalid answers.}
Invalid generated answers are report-only in all three political conditions; their rates are $6.25$--$9.38\%$ in Gemma~4 and zero in both Qwen models. All 128 prompts remain in the likelihood readout and generated-rate denominator. All nine students in each model pass the VQA and degeneration checks. \rev{On the corrected VQA probe,} Gemma~4's VQA accuracy is $.508$--$.547$ against a $.441$ baseline, Qwen3's is $.605$--$.625$ against a $.648$ baseline, and Qwen2.5's is $.621$--$.660$ against a $.586$ baseline. Empty, refusal, and decoding-limit rates are zero in all three conditions. Student training uses 8,000 pairs, three epochs, and 375 updates, with learning rates $2\times10^{-4}$ in both Qwen models and $10^{-4}$ in Gemma~4. LLaVA-1.5 was also run on this target with the same design, three curation blocks with one student seed per block and arm. Each of its nine students hit the decoding limit on 7 of 128 prompts ($5.47\%$), above the $5\%$ degeneration limit, so the condition is pipeline-invalid; its raw contrasts are provided in the supplementary report.

\subsection{Candidate-normalized likelihood and generated answers}
\label{app:normalized-readout}

For each model, prompt, and block, we normalize the eight answer-body
likelihoods from Equation~\ref{eq:preference-readout}:
\[
 p_S^{\mathcal A}(j)=
 \frac{\exp\ell_S(j,a^\star)}{\sum_{a\in\mathcal A}\exp\ell_S(j,a)}
 =\sigma\bigl(r_S(j)-\log 7\bigr).
\]
We average these per-prompt shares over the 128 prompts and three blocks. The normalization covers only the eight answer strings, excluding other continuations and answer terminators, and differs from the parsed target-answer frequency under greedy decoding. It summarizes the same likelihood measurements, not an independent replication or a prediction of generated-answer frequency.

All 21 analyzed conditions have positive mean target--neutral normalized shares, including all 63 block contrasts. Generated-answer differences are positive in 17 conditions, zero in three, and negative in one: Qwen2.5's third owl seed set. These summaries reuse data and controls, include student-seed repeats, and do not add independent curation replications or change the criterion outcomes. This post-hoc table covers these selected conditions, not the full training grid or the caption-relay experiments. For example, Qwen2.5's normalized camel share rises from $37.8\%$ to $40.0\%$, while its generated camel rate stays at $1.8\%$. An average share among fixed candidate strings does not determine how often the target wins token-by-token greedy decoding over the full vocabulary.

\begin{table}[!b]
  \centering\footnotesize
  \caption{Normalized target likelihood and generated target-answer rates from the same evaluations, as neutral $\to$ target percentages averaged over three blocks. Generated rates keep all 128 prompts per block, invalid answers included, in the denominator. Signs in the text use full precision.}

  \label{tab:normalized-readout}
  \setlength{\tabcolsep}{3pt}
  \begin{tabular}{@{}lllrr@{}}
    \toprule
    Family & Target & Student setting
      & \makecell{Normalized\\$N\to T$ (\%)}
      & \makecell{Generated\\$N\to T$ (\%)} \\
    \midrule
    Qwen2.5-VL & owl & Original & $2.8 \to 3.3$ & $6.3 \to 7.6$ \\
    Qwen2.5-VL & owl & Seed set 2 & $3.1 \to 3.7$ & $6.3 \to 8.1$ \\
    Qwen2.5-VL & owl & Seed set 3 & $3.2 \to 3.6$ & $6.5 \to 5.7$ \\
    Qwen3-VL & owl & Original & $24.0 \to 29.8$ & $28.6 \to 39.1$ \\
    Qwen3-VL & owl & Seed set 2 & $22.6 \to 28.9$ & $26.3 \to 34.9$ \\
    Qwen3-VL & owl & Seed set 3 & $22.8 \to 28.0$ & $25.5 \to 35.4$ \\
    Gemma~4 & owl & $10^{-4}$ & $16.2 \to 19.3$ & $2.6 \to 3.1$ \\
    Gemma~4 & owl & $5\!\times\!10^{-5}$ & $17.3 \to 21.1$ & $5.7 \to 7.6$ \\
    Gemma~4 & owl & $10^{-4}$, repeat 1 & $16.5 \to 20.0$ & $3.6 \to 4.9$ \\
    Gemma~4 & owl & $10^{-4}$, repeat 2 & $16.4 \to 20.3$ & $2.6 \to 3.9$ \\
    Qwen2.5-VL & camel & Original & $37.8 \to 40.0$ & $1.8 \to 1.8$ \\
    Qwen3-VL & camel & Original & $3.6 \to 5.3$ & $0.0 \to 0.3$ \\
    Gemma~4 & camel & Original & $0.1 \to 0.2$ & $0.0 \to 0.0$ \\
    Qwen2.5-VL & oak & Original & $23.4 \to 24.6$ & $6.8 \to 7.8$ \\
    Qwen3-VL & oak & Original & $52.4 \to 54.5$ & $85.7 \to 89.6$ \\
    Gemma~4 & oak & Original & $21.3 \to 27.1$ & $51.6 \to 59.4$ \\
    LLaVA-1.5 & owl & Original & $5.3 \to 5.6$ & $5.2 \to 6.3$ \\
    LLaVA-1.5 & camel & Original & $6.7 \to 7.2$ & $9.4 \to 9.9$ \\
    Gemma~4 & Reagan & Original & $2.4 \to 4.5$ & $0.0 \to 0.0$ \\
    Qwen3-VL & Reagan & Original & $7.4 \to 9.8$ & $4.2 \to 7.8$ \\
    Qwen2.5-VL & Reagan & Original & $5.0 \to 5.7$ & $1.3 \to 2.3$ \\
    \bottomrule
  \end{tabular}
\end{table}

\subsection{JPEG-quality diagnostics}
\label{app:jpeg-diagnostics}

Table~\ref{tab:jpeg-diagnostics} summarizes exploratory JPEG-quality
differences between owl-target and neutral selections. The negative shifts
in the two Qwen families motivated the matched-selection experiments in
Section~\ref{sec:selection-controls}. The Gemma~4 diagnostic was computed
subsequently, extending the earlier analysis to the fourth family.
LLaVA and Gemma~4 show smaller mean shifts in the opposite direction, and
neither family was tested with the matched training intervention.
These mean comparisons do not establish equality of the full quality
distributions or rule out compression-related contributions to transfer.

\begin{table}[htbp]
  \centering
  \small
  \caption{JPEG-quality differences between owl-target and neutral
  selections. Entries are paired means (SE) of quality$(T)-$quality$(N)$,
  restricted to sources for which the selected variants differ. Units are
  JPEG quality points on the 94--100 scale; negative values indicate lower
  target-selected quality. SEs describe variation across discordant sources
  within a block, not variation across training runs.}
  \label{tab:jpeg-diagnostics}
  \setlength{\tabcolsep}{5pt}
  \begin{tabular}{@{}lrrr@{}}
    \toprule
    Family & Block 1 & Block 2 & Block 3 \\
    \midrule
    LLaVA-1.5  & $+0.093\ (0.068)$ & $+0.197\ (0.066)$ & $+0.039\ (0.068)$ \\
    Qwen2.5-VL & $-0.323\ (0.057)$ & $-0.207\ (0.057)$ & $-0.327\ (0.057)$ \\
    Qwen3-VL   & $-0.657\ (0.047)$ & $-0.719\ (0.046)$ & $-0.617\ (0.046)$ \\
    Gemma~4    & $+0.022\ (0.045)$ & $+0.032\ (0.044)$ & $+0.034\ (0.045)$ \\
    \bottomrule
  \end{tabular}
\end{table}

The numbers of discordant sources in blocks 1/2/3 are
1,800/1,809/1,791 for LLaVA, 2,444/2,488/2,526 for Qwen2.5,
3,734/3,832/3,709 for Qwen3, and 3,889/4,049/3,928 for Gemma~4.

\subsection{Randomized selections with matched JPEG counts}
\label{app:matched-selection}
The randomized control constructs all three arms, including a second
neutral arm. Its criterion therefore uses its own matched neutral--neutral
sham, not the original cohort's sham. Table~\ref{tab:matched-selection-details}
reports the matched block contrasts and original-minus-matched differences.
All matched students pass eligibility, but neither matched condition meets
the preference criterion. These differences are descriptive comparisons of
one randomized assignment per arm and block, not an equivalence test or
proof that source-specific choice is the unique carrier.
\begin{table}[htbp]
\centering\small
\caption{JPEG-count-matched controls, nats. $E^M_b$ uses matched target
and matched neutral selections; $h_M$ uses the matched neutral and second
neutral selections. $R_b=E_b-E^M_b$ compares original and matched effects.}
\label{tab:matched-selection-details}
\setlength{\tabcolsep}{4pt}
\begin{tabular}{@{}lrrrrrrr@{}}
\toprule
Family & $E^M_1$ & $E^M_2$ & $E^M_3$ & $h_M$ & $R_1$ & $R_2$ & $R_3$ \\
\midrule
Qwen2.5-VL & -0.020 & 0.126 & 0.028 & 0.109 & 0.219 & 0.133 & 0.109 \\
Qwen3-VL & 0.058 & -0.118 & 0.016 & 0.091 & 0.390 & 0.437 & 0.304 \\
\bottomrule
\end{tabular}
\end{table}
\rev{\subsection{Count-matched random reselection (owl; all four models)}
\label{app:count-matched-random}
This control asks whether the original owl contrasts of
Qwen3-VL, Gemma~4, Qwen2.5-VL and LLaVA-1.5 exceed what count-matched random
reselection produces. For each curation block $b$ we train, with the main-experiment student recipe of the family,
students on selections in which as many source photographs as the original target--neutral disagreement count are drawn at random from the
8,000 and the variant is drawn from the seven non-neutral candidates
(1a), students on selections that keep the actual target--neutral
disagreement sources and draw the variant from the six candidates other
than the target and neutral choices (1b), and retrainings of the
neutral student (replay). Each random student is contrasted with the original neutral student of its block ($Z$, in nats, using the preference readout; LLaVA-1.5 uses the B200 references described below). The original $E_b$ is shown for comparison; its outcome is not re-judged. The count-matched reselection analysis is report-only: no verdict is assigned and the neutral--neutral sham $h$ is not replaced by a new null distribution. We first ran five draws per scheme and block. After inspecting
them, we fixed extensions to ten draws per block, first for Qwen3-VL and
Gemma~4 and then for Qwen2.5-VL and LLaVA-1.5, retaining all initial draws.
The final ten-draw results are given first, followed by the LLaVA-1.5
references and the initial five-draw analysis.

\begin{table}[htbp]
  \centering\small
  \caption{Owl count-matched random reselection over draws 1--10 (ten per
  scheme and block). $E_b$ is the target--neutral contrast; $Z$ ranges give
  the minimum and maximum over the ten draws (nats). The fifth column gives
  the largest 1a $|Z|$ and whether $E_b$ exceeds it. The last column counts
  draws with $Z<E_b$. For LLaVA-1.5 all values use the B200 reference cohort
  (original selections and student seeds retrained on B200); the original H200 contrasts are $E_b=.059$, $.098$, and $.111$.}
  \label{tab:count-matched-random-10}
  \setlength{\tabcolsep}{4pt}
  \begin{tabular}{@{}crccll@{}}
    \toprule
    $b$ & $E_b$ & 1a $Z$ range & 1b $Z$ range & 1a $\max|Z|$ & $Z<E_b$ (1a, 1b) \\
    \midrule
    \multicolumn{6}{@{}l}{Qwen3-VL} \\
    1 & .447 & $-.155$ to .105 & .059 to .192 & .155 (holds) & 10/10, 10/10 \\
    2 & .319 & $-.145$ to $-.002$ & $-.013$ to .129 & .145 (holds) & 10/10, 10/10 \\
    3 & .321 & .022 to .132 & .068 to .174 & .132 (holds) & 10/10, 10/10 \\
    \midrule
    \multicolumn{6}{@{}l}{Gemma~4 (original $10^{-4}$ cohort)} \\
    1 & .281 & $-.099$ to .136 & $-.134$ to .071 & .136 (holds) & 10/10, 10/10 \\
    2 & .392 & $-.056$ to .097 & $-.073$ to .155 & .097 (holds) & 10/10, 10/10 \\
    3 & .304 & $-.379$ to $-.097$ & $-.282$ to $-.068$ & .379 (misses) & 10/10, 10/10 \\
    \midrule
    \multicolumn{6}{@{}l}{Qwen2.5-VL} \\
    1 & .199 & .021 to .102 & $-.055$ to .108 & .102 (holds) & 10/10, 10/10 \\
    2 & .259 & .021 to .200 & $-.014$ to .161 & .200 (holds) & 10/10, 10/10 \\
    3 & .137 & $-.133$ to .076 & $-.149$ to .038 & .133 (holds) & 10/10, 10/10 \\
    \midrule
    \multicolumn{6}{@{}l}{LLaVA-1.5 (B200 reference cohort)} \\
    1 & .063 & $-.097$ to .041 & .024 to .100 & .097 (misses) & 10/10, 7/10 \\
    2 & .063 & $-.040$ to .047 & .009 to .074 & .047 (holds) & 10/10, 8/10 \\
    3 & .107 & $-.087$ to .046 & .013 to .066 & .087 (holds) & 10/10, 10/10 \\
    \bottomrule
  \end{tabular}
\end{table}

\paragraph{Final ten-draw results.} Table~\ref{tab:count-matched-random-10}
summarizes all ten draws and underlies Figure~\ref{fig:rn-signed-controls}.
\emph{Signed comparison:} $E_b$ exceeds every 1a and 1b $Z_{b,j}$ in Qwen3-VL,
Qwen2.5-VL, and Gemma~4 (60 of 60 students each). In LLaVA-1.5 it exceeds all
30 1a draws and 25 of 30 1b draws (7, 8, and 10 of 10 in blocks 1--3), so 235
of the 240 comparisons are positive. \emph{Magnitude markers:} the blockwise
marker, $E_b$ above the largest 1a $|Z_{b,j}|$, holds in every block of
Qwen3-VL and Qwen2.5-VL and in two of three blocks of Gemma~4 (block~3:
$\max|Z|=.379$ against $E_3=.304$) and LLaVA-1.5 (block~1: $\max|Z|=.097$
against $E_1=.063$); adding draws cannot recover a failed blockwise marker.
The mean marker holds in all four models: $\bar E$ is $.362$, $.198$, $.326$,
and $.078$ against $2\,\mathrm{mean}|Z|$ of $.133$, $.139$, $.251$, and $.061$
for Qwen3-VL, Qwen2.5-VL, Gemma~4, and LLaVA-1.5. Against the fixed
disagreement sources (1b), $E_b-\bar Z$ averages $.264$, $.178$, $.368$, and
$.031$ in the same order; LLaVA's block values are $.008$, $.017$, and $.068$.
\emph{Ordered pairs:} ordered pairs of 1a students, treated as synthetic
target/neutral cohorts with the reference neutral students, fire the transfer
criterion in 0 of 90 cases per model (20 within draws 1--5, 20 within draws
6--10, and 50 across them). All 132 students of the extensions pass the eligibility rule under the original and the corrected probe
(Appendix~\ref{app:corrected-vqa}), and the reference neutral students are
eligible, so every pair is evaluable. The status of these counts and of the
1b comparison is stated once, at the end of this subsection.

\paragraph{LLaVA-1.5 references.} LLaVA-1.5's original students were trained and evaluated on H200, whereas all of its random students were trained
on B200, so the two analyses use different B200 references
(Table~\ref{tab:count-matched-llava-references}). A specification fixed before
any reference student was trained retrained the original students of each block on B200 from the original selection manifests, with the original student seed and example order, and made this cohort's target and neutral
students the reference of the ten-draw comparison; the replacement did not
depend on the cohort's outcome. Retraining on B200 shifted the original students' readouts by between $-.053$ and $-.006$ nats; this difference
combines retraining, platform, and environment differences and is comparable
in size to LLaVA's effect, so the ten-draw comparison keeps target, neutral,
and random students in one training environment. Applied descriptively, the transfer criterion also holds for this reference cohort (positive block
contrasts, $\bar E\ge 2h$, owl-maximal profile, eligible students); the original H200 values and outcome are unchanged. The initial five-draw
analysis instead used a B200 re-evaluation of the H200-trained original students, with the re-evaluated neutral student as the common baseline. Its
neutral replays differ from that neutral student, whereas the reference
neutral student of the ten-draw analysis is bit-identical (adapter and
projector) to the neutral replays, so its replay contrast is zero in every
block; the two replay results are separate observations.

\begin{table}[htbp]
  \centering\footnotesize
  \caption{\rev{LLaVA-1.5 owl references of the count-matched random-reselection
  analyses. $E_b$, $\bar E$, and $h$ are target--neutral contrasts and the sham
  magnitude in nats. Replay is the retrained neutral student minus the
  reference neutral student in each block.}}
  \label{tab:count-matched-llava-references}
  \setlength{\tabcolsep}{3pt}
  \rev{\begin{tabular}{@{}>{\raggedright\arraybackslash}p{.25\linewidth}>{\raggedright\arraybackslash}p{.23\linewidth}rrr>{\raggedright\arraybackslash}p{.14\linewidth}@{}}
    \toprule
    Reference & Used in & \makecell[r]{$E_b$\\(blocks 1--3)} & $\bar E$ & $h$ & Replay \\
    \midrule
    Original H200 run & Table~\ref{tab:owl-readouts}; original outcome & .059, .098, .111 & .089 & .036 & -- \\
    B200 re-evaluation of the H200-trained students & Five-draw analysis (Table~\ref{tab:count-matched-random}) & .074, .096, .105 & .092 & .033 & $-.039$, $-.006$, $-.053$ \\
    B200 retrained reference cohort & Ten-draw analysis (Table~\ref{tab:count-matched-random-10}, Figure~\ref{fig:rn-signed-controls}) & .063, .063, .107 & .078 & .034 & 0, 0, 0 \\
    \bottomrule
  \end{tabular}}
\end{table}

\clearpage
\paragraph{Initial five-draw analysis.} Table~\ref{tab:count-matched-random}
lists the initial five draws as first analyzed, and
Table~\ref{tab:count-matched-random-5-summary} gives the markers of that
analysis for each model.

\begin{table}[htbp]
  \centering\small
  \caption{Owl count-matched random reselection, initial five draws per
  scheme as first analyzed (all four models). $E_b$ is the original target--neutral contrast; $Z$ are contrasts of random students against the original neutral student (nats). 1a draws
  sources and variants at random; 1b keeps the disagreement sources. For
  LLaVA-1.5, $E_b$ and every $Z$ use the B200 re-evaluation of the original neutral student as the common baseline; the original H200 record is
  $E_b=.059$, $.098$, and $.111$.}
  \label{tab:count-matched-random}
  \setlength{\tabcolsep}{3pt}
  \begin{tabular}{@{}crlrlr@{}}
    \toprule
    $b$ & $E_b$ & 1a $Z$ (five draws) & Mean & 1b $Z$ (five draws) & Mean \\
    \midrule
    \multicolumn{6}{@{}l}{Qwen3-VL} \\
    1 & .447 & $.011,\ -.148,\ -.026,\ .013,\ .040$ & $-.022$ & $.120,\ .192,\ .144,\ .110,\ .115$ & $.136$ \\
    2 & .319 & $-.041,\ -.145,\ -.049,\ -.118,\ -.002$ & $-.071$ & $-.013,\ .127,\ -.004,\ .064,\ .041$ & $.043$ \\
    3 & .321 & $.022,\ .132,\ .054,\ .029,\ .068$ & $.061$ & $.131,\ .092,\ .118,\ .115,\ .068$ & $.105$ \\
    \midrule
    \multicolumn{6}{@{}l}{Gemma~4 (original $10^{-4}$ cohort)} \\
    1 & .281 & $.136,\ .061,\ -.084,\ -.057,\ -.045$ & $.002$ & $.005,\ -.077,\ -.111,\ .020,\ .011$ & $-.030$ \\
    2 & .392 & $.061,\ .075,\ .045,\ -.038,\ .097$ & $.048$ & $.074,\ .091,\ .071,\ .021,\ .072$ & $.066$ \\
    3 & .304 & $-.326,\ -.321,\ -.152,\ -.214,\ -.157$ & $-.234$ & $-.068,\ -.147,\ -.245,\ -.104,\ -.282$ & $-.169$ \\
    \midrule
    \multicolumn{6}{@{}l}{Qwen2.5-VL} \\
    1 & .199 & $.031,\ .035,\ .077,\ .038,\ .025$ & $.041$ & $.015,\ .108,\ .038,\ .033,\ -.055$ & $.028$ \\
    2 & .259 & $.046,\ .095,\ .021,\ .052,\ .123$ & $.068$ & $.090,\ .022,\ .136,\ .127,\ .084$ & $.092$ \\
    3 & .137 & $-.058,\ -.001,\ -.112,\ -.051,\ -.050$ & $-.054$ & $.009,\ -.051,\ -.118,\ -.048,\ .028$ & $-.036$ \\
    \midrule
    \multicolumn{6}{@{}l}{LLaVA-1.5 (B200 baseline)} \\
    1 & .074 & $-.135,\ -.110,\ -.040,\ -.040,\ -.001$ & $-.065$ & $-.002,\ .019,\ .001,\ .005,\ .062$ & $.017$ \\
    2 & .096 & $-.020,\ .032,\ .007,\ .021,\ .025$ & $.013$ & $.025,\ .049,\ .067,\ .030,\ .058$ & $.046$ \\
    3 & .105 & $-.080,\ -.007,\ -.052,\ -.074,\ -.064$ & $-.055$ & $.000,\ .013,\ .001,\ -.016,\ -.005$ & $-.001$ \\
    \bottomrule
  \end{tabular}
\end{table}

\begin{table}[htbp]
  \centering\footnotesize
  \caption{\rev{Markers of the initial five-draw analysis
  (Table~\ref{tab:count-matched-random}), in nats. The blockwise marker compares
  $E_b$ with the largest 1a $|Z|$; the mean marker compares $\bar E$ with twice
  the mean 1a $|Z|$. LLaVA-1.5 uses the B200 re-evaluation baseline of
  Table~\ref{tab:count-matched-llava-references}.}}
  \label{tab:count-matched-random-5-summary}
  \setlength{\tabcolsep}{3pt}
  \rev{\begin{tabular}{@{}l>{\raggedright\arraybackslash}p{.25\linewidth}>{\raggedright\arraybackslash}p{.13\linewidth}>{\raggedright\arraybackslash}p{.15\linewidth}>{\raggedright\arraybackslash}p{.26\linewidth}@{}}
    \toprule
    Model & Blockwise marker & Mean marker & 1b: $E_b-\bar Z$ (mean) & Neutral replay \\
    \midrule
    Qwen3-VL & Holds in every block ($\max|Z|=.148$) & $.362$ vs $2\times.060=.120$ & .311, .276, .216 (.268) & Original neutral weights and stored preference and VQA outputs reproduced exactly; three configuration hashes differ from the original receipts \\
    Gemma~4 & Two of three blocks; in block~3 all five 1a draws are negative and $\max|Z|=.326$ exceeds $E_3=.304$ & $.326$ vs $.249$ & .312, .326, .473 (.370) & Original neutral weights and stored outputs reproduced exactly \\
    Qwen2.5-VL & Holds in every block ($\max|Z|=.123$) & $.198$ vs $.109$ & .171, .167, .173 (.170) & Original neutral weights and stored preference outputs reproduced exactly; configuration receipts differ \\
    LLaVA-1.5 & Two of three blocks; in block~1 all five 1a draws are negative and $\max|Z|=.135$ exceeds $E_1=.074$ & $.092$ vs $2\times.047=.094$, just short & .057, .050, .106 (.071) & Differs by $-.039$, $-.006$, $-.053$; combines retraining, platform, and environment differences, both sides evaluated on B200, and is not required to be zero \\
    \bottomrule
  \end{tabular}}
\end{table}

\paragraph{Common checks and status.} In each model the initial analysis has
33 students, all of which pass the eligibility rule (VQA at least
$U_{\mathrm{base}}-0.10$, degeneration at most $5\%$; invalid answers
report-only) under the original probe and remain eligible under the corrected
probe (Appendix~\ref{app:corrected-vqa}); for Qwen2.5-VL the original-probe
clean base is $U_{\mathrm{base}}=.582$. In Gemma~4, 9 of the 33 students have
invalid-answer rates between 5.5\% and 6.2\%, which are report-only under this
rule. The original neutral references also remain eligible under the corrected probe, so each model's twenty ordered pairs of 1a students, fixed before the analysis and treated as synthetic target/neutral cohorts with the original neutral students, stay evaluable; they fire the transfer criterion in 0 of 20
cases in every model. These counts, like the 0 of 90 of the ten-draw analysis,
are frequencies conditional on shared students, not independent tests or a
calibrated false-positive rate. In both analyses the 1b contrasts do not
identify the source set as the unique carrier, the 1a/1b difference is not
decomposed causally, and the comparisons are descriptive, not a calibrated
null test or a causal decomposition.
}

\rev{\subsection{Pair-preserving random assignment (Qwen3-VL owl)}
\label{app:pair-assignment}
This control keeps the original Qwen3-VL target and neutral picks and
randomizes only which arm receives each. At every source where the two picks
differ, an independent fair coin gives one picked variant to arm A and the
other to arm B; sources where they agree are identical in both arms.
Captions, sources, sample count, student recipe, seeds, and example order
follow the original Qwen3-VL owl condition. The original picks differ
on 3,734, 3,832, and 3,709 of the 8,000 sources in blocks 1--3, and arm A
received 1,835--1,936 target picks per assignment. Five assignments per block
were run first, and five more per block were fixed before the first five had
results (60 students in all). Each assignment gives a contrast
$\mu_A-\mu_B$ under the preference readout.

Averaged over blocks, the signed contrast is $.024$ nats in the first batch
(block means $.009$, $.033$, and $.031$) and $.010$ in the second ($.002$,
$.035$, and $-.007$). The largest absolute contrast per block is $.046$,
$.141$, and $.059$ in the first batch and $.137$, $.067$, and $.063$ in the
second; the original $E_b$ ($.447$, $.319$, $.321$) exceeds each, and
$\bar E=.362$ exceeds twice the mean absolute contrast in both batches
($2\times.042$ and $2\times.044$). Because the assignment is symmetric, a
signed mean near zero is expected by construction; the comparison places
the size of assignment variation next to the original contrast and is not
a permutation test, a calibrated null, or a mechanism. The ten assignments
per block are not independent curation replications. All 60 Qwen3-VL
students pass the eligibility rule under the original probe and remain eligible under the corrected probe; the report-only status is
unchanged.}

\subsection{Near-duplicate similarity and selection detectability}
\label{app:detectability}

We quantify visual similarity using L2-normalized CLIP ViT-L/14-336 image
embeddings~\citep{radford2021clip}. For each of the first 2,000 sources in
the shared candidate bank, we average cosine similarity over the 28 pairs
among its eight variants. The mean of these source-level averages is
$0.974$, and their minimum is $0.893$; the latter is not the minimum
individual-pair similarity. For sources on which two curator arms choose
different variants, the mean similarity between their choices ranges from
$0.974$ to $0.976$ across families, blocks, and arm pairs.

We also test whether a classifier can identify which of two curator arms
selected a variant. This is a balanced binary task, restricted to sources
where the arms disagree, rather than prediction of a candidate's index
among all eight variants. The features are either nine transformation
parameters (JPEG quality, brightness, contrast, red/blue gains, and four
crop coordinates) or the CLIP embedding. The single-variant probe classifies
individual selected variants. The paired probe instead trains on the
feature difference between the two selections and its negation, with
opposite labels. Both use logistic regression ($C=1$, at most 2,000
iterations) with standardization fitted only on each training fold.
Five-fold cross-validation groups both examples from a source in the same
fold: the single-variant probe uses shuffled stratified group folds
(seed 20260923), and the paired probe uses deterministic group folds. We report the amended source-grouped analysis;
the initial ungrouped analysis was excluded because it placed variants of
the same source across training and test folds.

\begin{table}[htbp]
  \centering
  \small
  \caption{Selecting-arm classification accuracy on discordant sources.
  Ranges span three blocks and all arm pairs: six pairs among the owl,
  camel, $N$, and $N'$ arms in LLaVA, and three pairs among $T,N,N'$ in
  each other model. Single classifies one selected variant; Paired classifies
  the difference between the two variants. Chance is $0.5$; accuracies are
  descriptive. Source counts are per block and arm pair. For Qwen3, the maximum source count of 3,834 is the
  target--second-neutral pair in block 3; the target--neutral counts are
  reported separately in Appendix~\ref{app:jpeg-diagnostics}.}
  \label{tab:detectability}
  \setlength{\tabcolsep}{4pt}
  \begin{tabular}{@{}lccccc@{}}
    \toprule
    Model & Sources & \multicolumn{2}{c}{Parameters}
      & \multicolumn{2}{c}{CLIP embeddings} \\
    \cmidrule(lr){3-4}\cmidrule(lr){5-6}
    & & Single & Paired & Single & Paired \\
    \midrule
    LLaVA-1.5   & 1{,}302--1{,}856 & 0.487--0.525 & 0.490--0.531 & 0.488--0.514 & 0.481--0.528 \\
    Qwen2.5-VL  & 2{,}444--3{,}736 & 0.497--0.531 & 0.494--0.547 & 0.500--0.525 & 0.501--0.533 \\
    Qwen3-VL    & 1{,}072--3{,}834 & 0.497--0.579 & 0.490--0.597 & 0.499--0.566 & 0.478--0.547 \\
    Gemma~4 E4B & 1{,}854--4{,}049 & 0.496--0.509 & 0.472--0.509 & 0.495--0.506 & 0.494--0.528 \\
    \bottomrule
  \end{tabular}
\end{table}

Across these probes, LLaVA and Gemma~4 accuracies remain close to $50\%$,
whereas Qwen3 reaches $59.7\%$ (Table~\ref{tab:detectability}).
Near-chance accuracy here does not establish statistical equivalence to
chance or rule out stronger detectors. These tests use linear classifiers,
not human raters or prompted LLM/VLM judges. They do not establish visual
imperceptibility or exclude a causal role for transformation parameters;
detectability and the matched-training intervention answer different
questions.

\subsection{Receiver-feature and initial-gradient probes}
\label{app:receiver-probes}

Exploratory probes use original LLaVA-1.5-7B-curated
images in three recipients at initialization
(Table~\ref{tab:receiver-probes}). For feature separability, we
mean-pool and L2-normalize visual tokens after the projector or merger,
center each T--N pair, and fit standardized logistic regression
($C=1$) with five folds grouped by source. We retain sources where both
T--N and N--N$'$ selections differ (752/739/731 sources per block).
T--N areas under the receiver operating characteristic curve (AUCs) span
.482--.549; the corresponding N--N$'$ sham AUCs span
.451--.523. These are receiver-feature AUCs, distinct from the CLIP
accuracies above; they do not establish indistinguishability.
For initial-gradient alignment, we sample 512 common sources per
block and compute $\mathbf{g}_{\mathrm{owl}}^\top\mathbf{d}_{TN}$, where
$\mathbf{g}_{\mathrm{owl}}$ is the gradient of the summed 128-prompt owl
readout and $\mathbf{d}_{TN}$ is the negative mean difference between
T- and N-image caption-loss gradients over trainable parameters.
Dropout is disabled. A positive value predicts a
local first-order increase under a small gradient step, not the full
training trajectory. Alignment is
positive in all blocks for LLaVA-1.5-13B despite no detected transfer
in that cross-checkpoint control
(Appendix~\ref{app:receiver-robustness}), so it is not sufficient to
explain transfer. These probes do not cover Qwen3-VL or Gemma~4.

\begin{table}[htbp]
  \centering
  \small
  \caption{Exploratory probes on LLaVA-1.5-7B-curated images. Feature AUC
  entries are mean held-out fold values; gradient-alignment signs refer
  to blocks 1/2/3.}
  \label{tab:receiver-probes}
  \begin{tabular}{@{}lrrrc@{}}
    \toprule
    Initial recipient & \makecell{Feature AUC\\Block 1}
      & \makecell{Feature AUC\\Block 2} & \makecell{Feature AUC\\Block 3}
      & \makecell{Gradient\\alignment} \\
    \midrule
    LLaVA-1.5-7B  & .530 & .486 & .488 & $+ / + / +$ \\
    LLaVA-1.5-13B & .549 & .482 & .501 & $+ / + / +$ \\
    Qwen2.5-VL-7B & .506 & .534 & .483 & $+ / - / -$ \\
    \bottomrule
  \end{tabular}
\end{table}

\subsection{Candidate breadth and curated-data fraction}
\label{app:breadth-mixture}
In separate LLaVA-1.5 development experiments, increasing the number of
available variants from $K=2$ to $8$ and $16$ increased the three-block mean
contrast from 0.036 to 0.105 and 0.126 nats, respectively
(Table~\ref{tab:breadth-mixture}). The $K=2$ cohort was positive in every
block but did not reach twice the sham magnitude. The $K=8$ comparison
uses the matched development cohort, rather than the confirmatory cohort
with $\bar E=0.089$. The increase from $K=8$ to $16$ was not uniform across
blocks.

A separate mixture experiment kept the 8,000 curated pairs and added
8,000 or 24,000 uncurated source-image--human-caption pairs, shared across
arms. With three epochs fixed, the 100\%, 50\%, and 25\% curated fractions
required 375, 750, and 1,500 updates, respectively. Mean contrasts decreased
from 0.109 to 0.050 and 0.046 nats; neither mixture met the transfer
criterion. These comparisons change dataset size and update count along
with the curated fraction, and do not establish a monotonic dose--response
law.

\begin{table}[htbp]
  \centering
  \small
  \setlength{\tabcolsep}{4pt}
  \caption{Candidate breadth and curated-data fraction in LLaVA-1.5
  development experiments. Block contrasts $E_b$, their mean $\bar E$,
  and sham magnitude $h$ are in nats. The two series use separate
  reference cohorts.}
  \label{tab:breadth-mixture}
  \begin{tabular}{@{}lrrrrrl@{}}
    \toprule
    Condition & $E_1$ & $E_2$ & $E_3$ & $\bar E$ & $h$ & Criterion \\
    \midrule
    Candidates: $K=2$ & 0.074 & 0.018 & 0.016 & 0.036 & 0.020 & Not met \\
    Candidates: $K=8$ & 0.121 & 0.131 & 0.063 & 0.105 & 0.028 & Met \\
    Candidates: $K=16$ & 0.103 & 0.157 & 0.117 & 0.126 & 0.020 & Met \\
    \midrule
    Curated fraction: 100\% & 0.142 & 0.063 & 0.122 & 0.109 & 0.017 & Met \\
    Curated fraction: 50\% & 0.098 & 0.056 & -0.002 & 0.050 & 0.020 & Not met \\
    Curated fraction: 25\% & -0.006 & 0.084 & 0.061 & 0.046 & 0.033 & Not met \\
    \bottomrule
  \end{tabular}
\end{table}

\section{Training-grid results}
\label{app:training-grid}

Tables~\ref{tab:training-llava}--\ref{tab:training-qwen3} give all completed
points in the receiver-training grid. Learning rates are in units of
$10^{-4}$; $E_1,E_2,E_3$, their mean $\bar E$, and sham $h$ are in nats.
The Eligible students column counts those passing the VQA and response-quality checks
out of nine. Met means that every requirement of
Section~\ref{sec:preference-readout} is satisfied. Not met means an
eligible condition fails a preference-transfer check, including conditions
with negative contrasts; it does not establish that an effect is absent.
Not evaluable means at least one student fails an eligibility check.
Its displayed contrasts are raw diagnostics and are never replaced by zero.
These are outcomes for individual settings, not necessarily the aggregate
conclusion of a multi-setting experiment. For eligible conditions, the last
column lists unmet checks: block sign means some $E_b\le0$, the sham floor
is $\bar E\ge2h$, and target profile means a positive maximum at the target.
An ineligible condition is marked eligibility without adjudicating its raw
preference contrasts.

The data sizes are nested subsets of a separately constructed 40K source
reservoir, rather than expansions of the original 8K existence cohort.
Budget and learning-rate comparisons reuse the selected manifests within
each model and data size. With an effective batch size of 64, nominal passes
are $64u/n$ for $u$ updates and $n$ pairs. Changing the update horizon also
changes warmup and cosine decay; changing source count at fixed updates
changes repetition. These comparisons do not isolate a pure repetition
effect or establish a family-specific optimum.

Each target--neutral comparison is within a host. Except for the
40,000-pair, 1,875-update condition at $8\times10^{-4}$, the Qwen3 grid
conditions use H200 for blocks 1--2 and B200 for block 3. That
$8\times10^{-4}$ condition uses B200 for all three blocks, so its comparison
with the corresponding $2\times10^{-4}$ and $4\times10^{-4}$ settings also
changes host allocation. \rev{No results are available for Qwen2.5's 40,000-pair experiments
at $4\times10^{-4}$ and $8\times10^{-4}$ in the reported grid.}

\begin{table}[htbp]
  \centering\small
  \caption{LLaVA-1.5 receiver-training grid. Outcomes and units are defined
  in Appendix~\ref{app:training-grid}. Values for not-evaluable conditions
  are raw diagnostics. Eligible students pass VQA and response-quality
  checks (out of nine).
  ``Met'' and ``Not met'' refer to the transfer criterion in
  Section~\ref{sec:preference-readout}. Eligibility counts and outcomes here use the original VQA evaluations; corrected-probe results are reported in Appendix~\ref{app:corrected-vqa}.}
  \label{tab:training-llava}
  \setlength{\tabcolsep}{3pt}
\begin{tabular}{@{}rrrrrrrrcl@{}}
\toprule
Pairs & Steps & \makecell{Learning rate\\($\times10^{-4}$)} & $E_1$ & $E_2$ & $E_3$ & $\bar E$ & $h$ & \makecell{Eligible\\students} & \makecell{Criterion /\\unmet checks} \\
\midrule
8K & 375 & 2 & 0.134 & 0.081 & 0.105 & 0.107 & 0.023 & 9/9 & Met \\
8K & 375 & 4 & 0.090 & 0.159 & 0.292 & 0.181 & 0.038 & 9/9 & Met \\
8K & 375 & 8 & 0.297 & 0.029 & 0.388 & 0.238 & 0.156 & 9/9 & \makecell[l]{Not met\\$\bar E<2h$} \\
16K & 375 & 2 & 0.047 & 0.044 & 0.079 & 0.057 & 0.019 & 9/9 & Met \\
16K & 375 & 4 & 0.027 & 0.128 & 0.092 & 0.082 & 0.027 & 9/9 & Met \\
16K & 375 & 8 & 0.201 & 0.015 & 0.183 & 0.133 & 0.059 & 9/9 & Met \\
16K & 750 & 2 & 0.236 & 0.129 & 0.153 & 0.172 & 0.011 & 9/9 & Met \\
16K & 750 & 4 & 0.273 & 0.187 & 0.047 & 0.169 & 0.121 & 9/9 & \makecell[l]{Not met\\$\bar E<2h$} \\
16K & 750 & 8 & -0.118 & -0.170 & 0.967 & 0.227 & 0.606 & 8/9 & \makecell[l]{Not evaluable\\eligibility} \\
16K & 1,500 & 2 & 0.415 & 0.309 & 0.191 & 0.305 & 0.059 & 9/9 & Met \\
40K & 1,500 & 2 & 0.194 & 0.120 & 0.163 & 0.159 & 0.048 & 9/9 & Met \\
40K & 1,875 & 2 & 0.263 & 0.062 & 0.226 & 0.184 & 0.093 & 9/9 & \makecell[l]{Not met\\$\bar E<2h$} \\
40K & 1,875 & 4 & -0.015 & -0.238 & -0.395 & -0.216 & 0.060 & 9/9 & \makecell[l]{Not met\\block sign\\$\bar G\le0$\\$\bar E<2h$\\target profile} \\
40K & 1,875 & 8 & 0.851 & -0.214 & 0.714 & 0.450 & 0.692 & 3/9 & \makecell[l]{Not evaluable\\eligibility} \\
\bottomrule
\end{tabular}
\end{table}

\begin{table}[htbp]
  \centering\small
  \caption{Qwen2.5-VL receiver-training grid. Completed points only;
  values for not-evaluable conditions are raw diagnostics. Eligible students
  pass VQA and response-quality checks (out of nine).
  ``Met'' and ``Not met'' refer to the transfer criterion in
  Section~\ref{sec:preference-readout}.}
  \label{tab:training-qwen25}
  \setlength{\tabcolsep}{3pt}
\begin{tabular}{@{}rrrrrrrrcl@{}}
\toprule
Pairs & Steps & \makecell{Learning rate\\($\times10^{-4}$)} & $E_1$ & $E_2$ & $E_3$ & $\bar E$ & $h$ & \makecell{Eligible\\students} & \makecell{Criterion /\\unmet checks} \\
\midrule
8K & 375 & 2 & 0.127 & 0.042 & 0.045 & 0.071 & 0.111 & 9/9 & \makecell[l]{Not met\\$\bar E<2h$} \\
8K & 375 & 4 & 0.234 & 0.140 & 0.023 & 0.132 & 0.160 & 9/9 & \makecell[l]{Not met\\$\bar E<2h$} \\
8K & 375 & 8 & -1.393 & -0.465 & -0.206 & -0.688 & 1.697 & 7/9 & \makecell[l]{Not evaluable\\eligibility} \\
16K & 375 & 2 & 0.038 & 0.018 & 0.109 & 0.055 & 0.037 & 9/9 & \makecell[l]{Not met\\$\bar E<2h$\\target profile} \\
16K & 375 & 4 & 0.125 & 0.002 & 0.017 & 0.048 & 0.081 & 9/9 & \makecell[l]{Not met\\$\bar E<2h$} \\
16K & 375 & 8 & -0.070 & 0.297 & -0.006 & 0.074 & 0.120 & 6/9 & \makecell[l]{Not evaluable\\eligibility} \\
16K & 750 & 2 & 0.180 & 0.172 & 0.104 & 0.152 & 0.045 & 9/9 & \makecell[l]{Not met\\target profile} \\
16K & 750 & 4 & 0.212 & 0.083 & 0.137 & 0.144 & 0.159 & 9/9 & \makecell[l]{Not met\\$\bar E<2h$\\target profile} \\
16K & 750 & 8 & 0.513 & -0.010 & -0.598 & -0.032 & 0.447 & 4/9 & \makecell[l]{Not evaluable\\eligibility} \\
16K & 1,500 & 2 & 0.247 & 0.622 & 0.295 & 0.388 & 0.124 & 9/9 & Met \\
40K & 1,500 & 2 & 0.149 & 0.133 & 0.184 & 0.155 & 0.085 & 9/9 & \makecell[l]{Not met\\$\bar E<2h$} \\
40K & 1,875 & 2 & 0.315 & 0.089 & 0.170 & 0.192 & 0.130 & 9/9 & \makecell[l]{Not met\\$\bar E<2h$} \\
\bottomrule
\end{tabular}
\end{table}

\begin{table}[htbp]
  \centering\small
  \caption{Qwen3-VL receiver-training grid. Values for not-evaluable conditions are raw diagnostics.
  The 40K, 1,875-update point at $8\times10^{-4}$ uses a different host
  allocation from the corresponding $2\times10^{-4}$ and $4\times10^{-4}$ points.
  Eligible students pass VQA and response-quality checks (out of nine).
  ``Met'' and ``Not met'' refer to the transfer criterion in
  Section~\ref{sec:preference-readout}. Eligibility counts and outcomes here use the original VQA evaluations; corrected-probe results are reported in Appendix~\ref{app:corrected-vqa}.}
  \label{tab:training-qwen3}
  \setlength{\tabcolsep}{3pt}
\begin{tabular}{@{}rrrrrrrrcl@{}}
\toprule
Pairs & Steps & \makecell{Learning rate\\($\times10^{-4}$)} & $E_1$ & $E_2$ & $E_3$ & $\bar E$ & $h$ & \makecell{Eligible\\students} & \makecell{Criterion /\\unmet checks} \\
\midrule
8K & 375 & 2 & 0.348 & 0.334 & 0.382 & 0.355 & 0.051 & 9/9 & Met \\
8K & 375 & 4 & 0.367 & 0.135 & 0.388 & 0.297 & 0.123 & 0/9 & \makecell[l]{Not evaluable\\eligibility} \\
8K & 375 & 8 & 0.152 & 0.113 & 0.664 & 0.310 & 0.255 & 0/9 & \makecell[l]{Not evaluable\\eligibility} \\
16K & 375 & 2 & 0.266 & 0.160 & 0.258 & 0.228 & 0.051 & 9/9 & Met \\
16K & 375 & 4 & 0.346 & 0.110 & 0.252 & 0.236 & 0.034 & 1/9 & \makecell[l]{Not evaluable\\eligibility} \\
16K & 375 & 8 & 0.203 & 0.127 & 0.507 & 0.279 & 0.171 & 0/9 & \makecell[l]{Not evaluable\\eligibility} \\
16K & 750 & 2 & 0.433 & 0.368 & 0.498 & 0.433 & 0.034 & 9/9 & Met \\
16K & 750 & 4 & 0.241 & 0.199 & 0.479 & 0.306 & 0.159 & 2/9 & \makecell[l]{Not evaluable\\eligibility} \\
16K & 750 & 8 & 0.990 & -0.172 & 1.124 & 0.648 & 0.427 & 0/9 & \makecell[l]{Not evaluable\\eligibility} \\
16K & 1,500 & 2 & 0.743 & 0.836 & 0.899 & 0.826 & 0.105 & 9/9 & Met \\
40K & 1,500 & 2 & 0.366 & 0.334 & 0.439 & 0.380 & 0.049 & 9/9 & Met \\
40K & 1,875 & 2 & 0.440 & 0.634 & 0.582 & 0.552 & 0.094 & 9/9 & Met \\
40K & 1,875 & 4 & 0.313 & 0.376 & 0.522 & 0.404 & 0.218 & 4/9 & \makecell[l]{Not evaluable\\eligibility} \\
40K & 1,875 & 8 & -0.110 & 0.942 & -0.036 & 0.265 & 0.289 & 0/9 & \makecell[l]{Not evaluable\\eligibility} \\
\bottomrule
\end{tabular}
\end{table}

For LLaVA at 40,000 pairs, 1,875 updates, and $8\times10^{-4}$, six
students fail VQA eligibility (accuracies $.488$--$.578$); one of those
also exceeds the degeneration-rate limit ($.094$). This is distinct from
the eligible negative contrast at $4\times10^{-4}$.

\subsection{Gemma~4 owl learning-rate follow-ups}
\label{app:gemma-owl-rates}
Table~\ref{tab:gemma-owl-rates} includes the originally attempted
$2\times10^{-4}$ condition and both lower-rate follow-ups on the same
selected data. Five of nine original students pass eligibility; the four
failures are in blocks 2 and 3. The original row is a raw diagnostic, not
evidence of detected transfer. VQA ranges are $.234$--$.445$ at
$2\times10^{-4}$ on the original probe, and $.500$--$.527$ at $10^{-4}$
and $.539$--$.559$ at $5\times10^{-5}$ on the corrected probe. Invalid-answer rates are report-only in these conditions.

\begin{table}[htbp]
\centering\small
\caption{Gemma~4 owl learning-rate conditions, 8,000 pairs and 375 updates.
Learning rates are in units of $10^{-4}$; contrasts and sham are nats.
The first row fails eligibility and retains its raw readout.}
\label{tab:gemma-owl-rates}
\setlength{\tabcolsep}{4pt}
\begin{tabular}{@{}rrrrrrcl@{}}
\toprule
Rate & $E_1$ & $E_2$ & $E_3$ & $\bar E$ & $h$ & Eligible & Criterion \\
\midrule
2 & 0.018 & 0.164 & 0.227 & 0.136 & 0.178 & 5/9 & Not evaluable \\
1 & 0.281 & 0.392 & 0.304 & 0.326 & 0.100 & 9/9 & Met \\
0.5 & 0.295 & 0.422 & 0.404 & 0.373 & 0.065 & 9/9 & Met \\
\bottomrule
\end{tabular}
\end{table}

\subsection{Scope of the utility screen}
\label{app:utility-scope}

The utility screen combines relative VQA accuracy with condition-specific
response-quality checks. Appendix~\ref{app:corrected-vqa} specifies the
VQA results and probe version used for each group of conditions.
Passing this relative screen does not establish preservation of general
visual ability. Preference-readout sensitivity is assessed separately in
Appendix~\ref{app:additional-sensitivity}.

\rev{\subsection{Learning trajectories}
\label{app:trajectories}
To see when the contrast appears during student training, we re-ran the
8K owl recipe (8,000 pairs, 375 updates) of all four models at three
learning rates each, saving checkpoints at updates 25, 50, 100, 200, and
375 and keeping the original selections, seeds, and example order. In
every block a control student trained without checkpoint saving has final
weights identical to its instrumented counterpart. At the main learning
rate the final readouts equal the original ones (for LLaVA-1.5, the B200
reference cohort of Appendix~\ref{app:count-matched-random}).
Table~\ref{tab:trajectories} lists the pooled contrasts. In all 12
model--rate settings the pooled contrast is positive at every saved
checkpoint from update 25 onward, so it appears within the first 25 of 375
updates; this does not locate the onset more precisely. At update 25, 35 of
36 block contrasts are positive (Qwen2.5-VL block~3 at $2\times10^{-4}$ is
$-.003$). The curves do not all rise monotonically: Qwen2.5-VL at
$4\times10^{-4}$ dips at updates 50 and 100, Gemma~4 at $10^{-4}$ dips at
update 200, and Gemma~4 at $2\times10^{-4}$ peaks at update 100 and then
declines. A higher learning rate does not always give a larger contrast.
Qwen3-VL at $4\times10^{-4}$ and Gemma~4 at $2\times10^{-4}$ fail the
original-probe utility check at some points, so their values are
diagnostics. These trajectories are report-only and assign no transfer
outcome.}

\begin{table}[htbp]
  \centering\small
  \caption{Pooled target--neutral contrast $\bar E$ (nats) at saved
  checkpoints of the 8K owl recipe (375 updates). Learning rates are in units
  of $10^{-4}$; $\ast$ marks each model's main rate, whose final value equals the original readout (for LLaVA-1.5, the B200 reference cohort; the original H200 value is .089). Utility counts original-probe passes over three
  blocks, five checkpoints, and three arms. Report-only.}
  \label{tab:trajectories}
  \setlength{\tabcolsep}{4pt}
  \begin{tabular}{@{}llrrrrrr@{}}
    \toprule
    Model & Rate & 25 & 50 & 100 & 200 & 375 & Utility \\
    \midrule
    Qwen3-VL    & 1         & .047 & .104 & .148 & .257 & .309 & 45/45 \\
                & 2$^\ast$  & .075 & .120 & .169 & .265 & .362 & 45/45 \\
                & 4         & .102 & .145 & .205 & .260 & .440 & 27/45 \\
    \addlinespace[2pt]
    Qwen2.5-VL  & 1         & .008 & .012 & .055 & .066 & .102 & 45/45 \\
                & 2$^\ast$  & .014 & .034 & .086 & .149 & .198 & 45/45 \\
                & 4         & .038 & .032 & .024 & .173 & .279 & 45/45 \\
    \addlinespace[2pt]
    Gemma~4     & 0.5       & .124 & .222 & .290 & .363 & .373 & 45/45 \\
                & 1$^\ast$  & .143 & .248 & .310 & .302 & .326 & 45/45 \\
                & 2         & .082 & .226 & .271 & .167 & .136 & 32/45 \\
    \addlinespace[2pt]
    LLaVA-1.5   & 1         & .016 & .018 & .022 & .042 & .046 & 45/45 \\
                & 2$^\ast$  & .018 & .030 & .033 & .073 & .078 & 45/45 \\
                & 4         & .021 & .045 & .066 & .085 & .151 & 45/45 \\
    \bottomrule
  \end{tabular}
\end{table}

\rev{We also re-ran LLaVA-1.5's 40K-pair, 1,875-update conditions at
$2\times10^{-4}$ and $4\times10^{-4}$ on B200, saving checkpoints from
update 10 to 1,875 (Table~\ref{tab:lltrajx}). This is a record of the
current run, not a reconstruction of the original runs' trajectories. At
$2\times10^{-4}$ the pooled contrast is positive from update 15 onward, dips
at updates 25 and 750, and ends at .176 (original .184), with all three
block endpoints positive. At $4\times10^{-4}$ it peaks at .072 at update 375,
changes sign twice, and ends at .022. Its block endpoints are $-.008$,
$+.471$, and $-.395$, so the pooled value averages blocks of opposite sign
and does not reproduce the original negative endpoint ($-.216$; blocks
$-.015$, $-.238$, $-.395$). The original references were trained on H200
for blocks 1--2 and on B200 for block~3; the block-3 adapters match their
references exactly and those of blocks 1--2 do not, and we do not attribute
this difference to a specific cause. All 21 final checkpoints pass the
corrected-probe eligibility rule against a corrected base of .6875
(minimum margin .0375). These runs are report-only.}

\begin{table}[htbp]
  \centering\small
  \caption{LLaVA-1.5 long-condition re-run (40K pairs, 1,875 updates, B200):
  pooled contrast $\bar E$ (nats) at saved checkpoints, block endpoints, and
  the original endpoints. Record of the current run; report-only.}
  \label{tab:lltrajx}
  \setlength{\tabcolsep}{6pt}
  \begin{tabular}{@{}lrr@{}}
    \toprule
    Update & $2\times10^{-4}$ & $4\times10^{-4}$ \\
    \midrule
    10    & $-.007$ & .004 \\
    15    & .003    & .007 \\
    25    & .001    & .003 \\
    50    & .016    & .012 \\
    100   & .028    & .033 \\
    200   & .050    & .028 \\
    375   & .088    & .072 \\
    750   & .078    & $-.028$ \\
    1,125 & .153    & .007 \\
    1,500 & .163    & .048 \\
    1,875 & .176    & .022 \\
    \midrule
    Block endpoints, re-run     & .204, .099, .226 & $-.008$, .471, $-.395$ \\
    Original $\bar E$ (blocks) & .184 (.263, .062, .226) & $-.216$ ($-.015$, $-.238$, $-.395$) \\
    \bottomrule
  \end{tabular}
\end{table}

\section{Caption-relay results and controls}
\label{app:caption-relay}

The original selection-trained students supply captions for a shared bank
of 7,999 images after one exact-byte training overlap was excluded. The
same fixed bank variant ($k=0$) is used across caption arms. Each block uses
its three first-generation students, one per arm; Gemma~4 uses the students
trained at $10^{-4}$. Their captions train fresh students.
A separate comparison uses one original curator per arm (the first of
the four trained curators), with the same generation settings. Changing the
receiver learning rate or seed reuses the saved captions and training
manifests rather than regenerating them.

The LLaVA $4\times10^{-4}$ condition was fixed before any parent relay student outcomes were opened. Its rate was informed by an
earlier receiver-training readout, not chosen after the parent relay's
failure. It is a development condition, not an independent confirmatory test or evidence of an optimal relay rate.

\paragraph{Caption generation.}
Every producer receives the image and the instruction
``Describe this image in one concise sentence.''
We use the checkpoint's chat serialization, greedy decoding
(\texttt{do\_sample=False}, \texttt{num\_beams=1}), and at most 48 new tokens,
with batches of 16, left padding, bfloat16, and scaled dot-product attention.
Temperature, top-$p$, and top-$k$ sampling are inactive. Other generation
settings are inherited from the checkpoint; Table~\ref{tab:caption-generation-settings}
records the repetition penalties and stopping tokens. Generation stops at
the first occurrence of any listed stop token or the length limit. We remove stop and padding
tokens, decode with special tokens skipped, and strip surrounding whitespace.
The resulting text is retained without further punctuation or Unicode
normalization.

The LLaVA system message is ``A chat between a curious human and an
artificial intelligence assistant. The assistant gives helpful, detailed,
and polite answers to the human's questions.'' The Qwen2.5 template inserts
``You are a helpful assistant.'' Qwen3 and Gemma~4 use no system message;
Gemma~4's prompt ends at its model-turn prefix without a thinking block.
No target-specific instruction is added. Producer LoRA weights are merged
into the matching base model for generation in evaluation mode; LLaVA also
restores each first-generation student's trained projector. The LLaVA
curators retain the base projector.

\begin{table}[htbp]
  \centering\small
  \caption{Caption-generation settings, shared across arms and all three
  blocks within each family. Repetition penalties are inherited, not tuned
  for the relay; LLaVA and Gemma~4 use the default with no repetition penalty
  (equivalent to 1.0). Stop and padding entries are tokenizer IDs.}
  \label{tab:caption-generation-settings}
  \setlength{\tabcolsep}{4pt}
  \begin{tabular}{@{}llrcll@{}}
    \toprule
    Family & Host & \makecell{Image\\side}
      & \makecell{Repetition\\penalty} & Stop IDs & Pad ID \\
    \midrule
    LLaVA-1.5  & B200 & 336 & 1.0  & 2 & 32001 \\
    Qwen2.5-VL & H200 & 336 & 1.05 & 151645 & 151643 \\
    Qwen3-VL   & B200 & 352 & 1.0  & 151645 & 151643 \\
    Gemma~4    & B200 & 768 & 1.0  & 106, 1, 50 & 0 \\
    \bottomrule
  \end{tabular}
\end{table}

Images are converted to RGB. LLaVA uses its checkpoint's CLIP processor.
Both Qwen models first resize the shortest side to 336 pixels with bicubic
interpolation and take a central $336\times336$ crop; the Qwen3 processor
then maps this to $352\times352$. Gemma~4 uses the corresponding 768-pixel
resize and central crop. These transformations are identical across arms
within each family.

\begin{samepage}
\paragraph{Filtering and matched source sets.}
The caption filter uses the following Python regular expression with
case-insensitive matching (\texttt{re.I}):
\par\smallskip
{\footnotesize\noindent\verb!\b(camels?|fox(?:es)?|otters?|owls?|pandas?|pigs?|tigers?|whales?)\b!\par}
\smallskip
\end{samepage}
Thus it matches the eight animal names and the listed plurals at word
boundaries. A caption also fails if it is empty after stripping whitespace,
or its recorded finish reason is neither a stop token nor the 48-token
limit. Length-capped captions are retained unless they otherwise fail these
checks. There is no additional refusal-phrase or sentence-count filter.
Broader terms such as ``bird,'' ``animal,'' and ``feather'' are recorded
but do not cause removal. These caption-retention rules are separate from
the generated preference-answer eligibility checks.

If any of the three first-generation student captions fails, we remove that
source from all three caption arms and from the student image-selection
comparison, preserving the common source order. For captions from curators,
we start with this retained set and additionally remove a source if any of
the three curator captions fails the same filter. Those exclusions do not
alter the source sets used with student captions or student image selection.
The protocol stops a block if filtering student captions removes more than
half the bank; an additional removal of more than half the retained set
excludes only the comparison using curator captions. Neither threshold was
reached.
A later LLaVA audit corrected its treatment of length-capped outputs to
reproduce this existing rule; it did not change captions, retained datasets,
or trained students.

The retained source counts by block are 7,990/7,991/7,990 in LLaVA,
7,988/7,988/7,989 in Qwen3, and 7,990/7,990/7,990 in Qwen2.5. Across the three
caption arms, the corresponding training corpora contain 71,913, 71,895,
and 71,910 rows. Fixed-lexicon audits found no owl-family word matches in
these retained corpora. The LLaVA raw-generation audit additionally covers
71,991 rows before joint filtering. These checks concern explicit word forms;
they neither certify human indistinguishability nor exclude semantic cues.
Gemma~4 retains 7,987/7,988/7,987 sources for student captions and student
image selection. Captions from curators retain
7,982/7,982/7,980 in LLaVA, 7,947/7,935/7,944 in Qwen3,
7,984/7,985/7,984 in Qwen2.5, and 7,857/7,848/7,849 in Gemma~4.

Tables~\ref{tab:relay-all-llava}--\ref{tab:relay-all-qwen25} group results by how the training pairs are obtained: captions from first-generation students, image selection by those students with human captions, or captions from the original curators. Each route has its own target and neutral arms. At $2\times10^{-4}$ and 375 updates, contrasts are larger with curator captions than with student captions in LLaVA and both Qwen models. The Qwen2.5 student-caption condition is ineligible, so its raw magnitude is descriptive. Differences in producer weights, filtering, and supervision prevent a causal ordering of curator authority. Student image selection is eligible but does not meet the transfer criterion in LLaVA and Qwen2.5; the Qwen3 and Gemma~4 conditions are not evaluable. Thus no tested second round of image selection supplies a criterion-level onward-transfer detection. Gemma~4's condition using curator captions is also not evaluable. Table~\ref{tab:relay-generated-rates} reports generated owl-answer rates at selected relay settings.

Original and Repeat distinguish the training/order seeds within a setting;
Repeat 1 and Repeat 2 are the two later Qwen2.5 runs at $10^{-4}$.
Learning rates are in units of $10^{-4}$; contrasts and sham magnitudes are
in nats. Met means that every transfer and eligibility requirement in
Section~\ref{sec:preference-readout} is satisfied. Not met means an
eligible condition fails a preference-transfer check. Not evaluable means
an eligibility failure, with the contrast retained only as a raw diagnostic.
Report only marks a setting designated for numerical reporting without a
transfer verdict; a dagger additionally marks failed eligibility.
A positive raw or report-only contrast is not relabeled as detected
transfer. The Eligible students column counts those passing the VQA and
response-quality checks out of nine\rev{: eligible second-generation students pass the VQA and degeneration checks and have invalid-answer rates of at most $5\%$}. For eligible conditions, peak names
the maximum of the pooled centered profile; for report-only settings it
is descriptive and does not assign a verdict. Listed unmet checks use
the definitions above, and an ineligible row is marked eligibility.

The LLaVA repeat at $4\times10^{-4}$ has the outcome Not replicated. Qwen3's repeat at $4\times10^{-4}$ meets the criterion, but
its original condition at that rate does not; its replication descriptor is Not applicable. \rev{At the primary $2\times10^{-4}$ recipe, Qwen3 also meets the transfer criterion with a new student seed set trained on the same captions ($\bar E=.806$, $h=.171$).} The Qwen2.5 follow-up at $10^{-4}$ was specified after
the parent utility failure, and both later seed sets reuse the same captions.
All these results are development evidence. A later Gemma~4 experiment at $10^{-4}$ and 375 updates was ineligible for all three
routes because some students exceeded its 5\% invalid-answer ceiling.
The numbers of eligible students were 3/9 for student captions, 6/9 for
student image selection, and 3/9 for curator captions. These conditions
are not evaluable; they are neither zero effects nor detected transfer.
Passing the VQA component alone would not satisfy the full eligibility rule.

\begin{table}[htbp]
  \centering\small
  \caption{LLaVA-1.5 caption relay and comparison routes. Outcome labels,
  seeds, and units are defined in Appendix~\ref{app:caption-relay}.
  Eligible students pass both VQA and response-quality checks (out of nine).
  ``Met'' and ``Not met'' refer to the transfer criterion in
  Section~\ref{sec:preference-readout}.}
  \label{tab:relay-all-llava}
  \setlength{\tabcolsep}{2.5pt}
\begin{tabular}{@{}rrrrrrrlcl@{}}
\toprule
\makecell{Learning rate\\($\times10^{-4}$)} & Steps & $E_1$ & $E_2$ & $E_3$ & $\bar E$ & $h$ & \makecell{Student\\seeds} & \makecell{Eligible\\students} & \makecell{Criterion /\\checks and peak} \\
\midrule
\multicolumn{10}{@{}l}{\textit{Captions from students}} \\
2 & 375 & 0.456 & -0.143 & 0.230 & 0.181 & 0.200 & Original & 9/9 & \makecell[l]{Not met\\block sign\\$\bar E<2h$\\peak: owl} \\
4 & 375 & 0.255 & 0.103 & 0.526 & 0.295 & 0.137 & Original & 9/9 & \makecell[l]{Met\\peak: owl} \\
8 & 375 & -0.173 & -0.245 & 0.262 & -0.052 & 0.606 & Original & 9/9 & \makecell[l]{Report only\\peak: camel} \\
4 & 375 & 0.367 & -0.041 & 0.362 & 0.229 & 0.121 & Repeat & 9/9 & \makecell[l]{Not met\\block sign\\$\bar E<2h$\\peak: owl} \\
3 & 375 & 0.301 & -0.219 & 0.390 & 0.157 & 0.192 & Original & 9/9 & \makecell[l]{Report only\\peak: owl} \\
6 & 375 & 0.854 & -0.266 & 0.197 & 0.262 & 0.233 & Original & 9/9 & \makecell[l]{Report only\\peak: owl} \\
4 & 750 & 0.692 & -0.375 & 0.194 & 0.170 & 0.384 & Original & 9/9 & \makecell[l]{Report only\\peak: owl} \\
\addlinespace[4pt]
\multicolumn{10}{@{}l}{\textit{Image selection by students}} \\
2 & 375 & -0.023 & 0.000 & 0.026 & 0.001 & 0.031 & Original & 9/9 & \makecell[l]{Not met\\block sign\\$\bar E<2h$\\target profile\\peak: pig} \\
\addlinespace[4pt]
\multicolumn{10}{@{}l}{\textit{Captions from curators}} \\
2 & 375 & 1.566 & 1.449 & 1.653 & 1.556 & 0.226 & Original & 9/9 & \makecell[l]{Met\\peak: owl} \\
\bottomrule
\end{tabular}
\end{table}

\begin{table}[htbp]
  \centering\small
  \caption{Qwen3-VL caption relay and comparison routes. \rev{Student-caption repeats at $2\times10^{-4}$ and $4\times10^{-4}$ retrain on the same captions with new training and order seeds.} Eligible students pass \rev{corrected-probe} VQA and response-quality
  checks (out of nine). $\dagger$: report-only and ineligible.
  ``Met'' and ``Not met'' refer to the transfer criterion in
  Section~\ref{sec:preference-readout}.}
  \label{tab:relay-all-qwen3}
  \setlength{\tabcolsep}{2.5pt}
\begin{tabular}{@{}rrrrrrrlcl@{}}
\toprule
\makecell{Learning rate\\($\times10^{-4}$)} & Steps & $E_1$ & $E_2$ & $E_3$ & $\bar E$ & $h$ & \makecell{Student\\seeds} & \makecell{Eligible\\students} & \makecell{Criterion /\\checks and peak} \\
\midrule
\multicolumn{10}{@{}l}{\textit{Captions from students}} \\
2 & 375 & 0.809 & 0.294 & 1.062 & 0.722 & 0.207 & Original & 9/9 & \makecell[l]{Met\\peak: owl} \\
\rev{2} & \rev{375} & \rev{0.987} & \rev{0.485} & \rev{0.946} & \rev{0.806} & \rev{0.171} & \rev{Repeat} & \rev{9/9} & \rev{\makecell[l]{Met\\peak: owl}} \\
4 & 375 & 0.268 & 0.040 & 1.357 & 0.555 & 0.542 & Original & 9/9 & \makecell[l]{Not met\\$\bar E<2h$\\target profile\\peak: fox} \\
4 & 375 & 0.555 & 0.727 & 0.611 & 0.631 & 0.173 & Repeat & 9/9 & \makecell[l]{Met\\peak: owl} \\
3 & 375 & 0.636 & 0.370 & 1.461 & 0.822 & 0.111 & Original & 9/9 & \makecell[l]{Report only\\peak: owl} \\
6 & 375 & 0.123 & 0.588 & 0.914 & 0.541 & 0.285 & Original & 9/9 & \makecell[l]{Report only\\peak: camel} \\
8 & 375 & -0.095 & 0.813 & -0.008 & 0.237 & 0.592 & Original & \rev{6/9} & \makecell[l]{Report only$^{\dagger}$\\eligibility} \\
4 & 750 & 0.986 & -0.278 & 2.024 & 0.911 & 0.709 & Original & 9/9 & \makecell[l]{Report only\\peak: owl} \\
\addlinespace[4pt]
\multicolumn{10}{@{}l}{\textit{Image selection by students}} \\
2 & 375 & 0.287 & 0.053 & 0.030 & 0.123 & 0.115 & Original & 7/9 & \makecell[l]{Not evaluable\\eligibility} \\
\addlinespace[4pt]
\multicolumn{10}{@{}l}{\textit{Captions from curators}} \\
2 & 375 & 21.962 & 21.142 & 23.293 & 22.132 & 0.160 & Original & 9/9 & \makecell[l]{Met\\peak: owl} \\
\bottomrule
\end{tabular}
\end{table}

\begin{table}[htbp]
  \centering\small
  \caption{Qwen2.5-VL caption relay and comparison routes. Student-caption
  runs at $10^{-4}$, including the two repeats, pass eligibility but do not
  meet the preference-transfer criterion. Eligible students pass VQA and
  response-quality checks (out of nine). $\dagger$: report-only and ineligible.
  ``Met'' and ``Not met'' refer to the transfer criterion in
  Section~\ref{sec:preference-readout}. Eligibility counts and outcomes here use the original VQA evaluations; corrected-probe results are reported in Appendix~\ref{app:corrected-vqa}.}
  \label{tab:relay-all-qwen25}
  \setlength{\tabcolsep}{2.5pt}
\begin{tabular}{@{}rrrrrrrlcl@{}}
\toprule
\makecell{Learning rate\\($\times10^{-4}$)} & Steps & $E_1$ & $E_2$ & $E_3$ & $\bar E$ & $h$ & \makecell{Student\\seeds} & \makecell{Eligible\\students} & \makecell{Criterion /\\checks and peak} \\
\midrule
\multicolumn{10}{@{}l}{\textit{Captions from students}} \\
2 & 375 & -0.179 & 0.144 & 0.498 & 0.154 & 0.180 & Original & 6/9 & \makecell[l]{Not evaluable\\eligibility} \\
4 & 375 & -0.052 & 0.739 & 0.660 & 0.449 & 0.254 & Original & 0/9 & \makecell[l]{Not evaluable\\eligibility} \\
4 & 375 & -0.050 & 0.288 & 0.544 & 0.261 & 0.169 & Repeat & 0/9 & \makecell[l]{Not evaluable\\eligibility} \\
3 & 375 & -0.184 & 0.420 & 0.496 & 0.244 & 0.399 & Original & 0/9 & \makecell[l]{Report only$^{\dagger}$\\eligibility} \\
6 & 375 & 0.306 & 1.123 & 0.662 & 0.697 & 0.315 & Original & 0/9 & \makecell[l]{Report only$^{\dagger}$\\eligibility} \\
8 & 375 & 0.953 & 0.433 & -0.589 & 0.266 & 0.781 & Original & 0/9 & \makecell[l]{Report only$^{\dagger}$\\eligibility} \\
4 & 750 & -0.641 & -0.078 & 0.946 & 0.076 & 0.760 & Original & 0/9 & \makecell[l]{Report only$^{\dagger}$\\eligibility} \\
1 & 375 & -0.095 & 0.200 & 0.303 & 0.136 & 0.103 & Original & 9/9 & \makecell[l]{Not met\\block sign\\$\bar E<2h$\\peak: owl} \\
1 & 375 & -0.110 & 0.149 & 0.267 & 0.102 & 0.072 & Repeat 1 & 9/9 & \makecell[l]{Not met\\block sign\\$\bar E<2h$\\peak: owl} \\
1 & 375 & -0.039 & 0.187 & 0.249 & 0.132 & 0.121 & Repeat 2 & 9/9 & \makecell[l]{Not met\\block sign\\$\bar E<2h$\\peak: owl} \\
\addlinespace[4pt]
\multicolumn{10}{@{}l}{\textit{Image selection by students}} \\
2 & 375 & -0.027 & 0.087 & 0.112 & 0.058 & 0.127 & Original & 9/9 & \makecell[l]{Not met\\block sign\\$\bar E<2h$\\peak: owl} \\
\addlinespace[4pt]
\multicolumn{10}{@{}l}{\textit{Captions from curators}} \\
2 & 375 & 3.234 & 3.641 & 3.509 & 3.461 & 0.332 & Original & 9/9 & \makecell[l]{Met\\peak: owl} \\
\bottomrule
\end{tabular}
\end{table}

\begin{table}[htbp]
  \centering\small
  \caption{Generated owl-answer rates at selected original relay settings.
  Rates aggregate 128 prompts across three blocks (384 responses per arm,
  including invalid outputs in the denominator). Differences use unrounded
  rates. Learning rates are in units of $10^{-4}$; $\bar E$ is in nats.
  These frequencies
  complement the likelihood readout and are not an independent set of runs.}
  \label{tab:relay-generated-rates}
  \setlength{\tabcolsep}{5pt}
\begin{tabular}{@{}lrrrrr@{}}
\toprule
Family & \makecell{Learning rate\\($\times10^{-4}$)} & $\bar E$ & Owl: $T$ & Owl: $N$ & $\Delta$ (pp) \\
\midrule
LLaVA-1.5 & 4 & 0.295 & 7.8\% & 7.6\% & $+0.3$ \\
Qwen3-VL & 2 & 0.722 & 37.2\% & 30.2\% & $+7.0$ \\
Qwen2.5-VL & 1 & 0.136 & 15.1\% & 14.6\% & $+0.5$ \\
\bottomrule
\end{tabular}
\end{table}

\clearpage
\section{Readout sensitivity and VQA utility}
\label{app:additional-sensitivity}
\label{app:llava-additional-sensitivity}

The preference-readout sensitivity analyses reuse fixed checkpoints from
the original LLaVA cohorts and the main Qwen2.5-VL, Qwen3-VL, and Gemma~4 experiments.
They assess sensitivity of the recorded measurements and do not add
independent curator, student-training, or source-sample replications.
Within each condition, block repeats hold the source and candidate-image
bank fixed; student-seed repeats assess training variability conditional on the selected data, while
the bootstrap assesses prompt variation conditional on the trained models.
None of these block repeats, student-seed repeats, or prompt bootstraps
samples new source images.
The LLaVA analyses and the Qwen/Gemma extension are reported separately,
followed by the VQA utility results.

\subsection{LLaVA preference prompts and conditional uncertainty}

On a second, non-identical 128-prompt bank with the same answer set, the
owl and camel contrasts remain positive in each of the three blocks
(Table~\ref{tab:llava-readout-sensitivity}). An additional diagnostic identified
an ambiguous question family: a question about learning a trick can elicit
a trick rather than an animal. Excluding its eight answer-order rotations
leaves 120 prompts. Excluding that family and a second family about a jam
label leaves 112 prompts. The contrasts remain positive in all three blocks
for both targets under each of these subsets.

\begin{table}[htbp]
  \centering\small
  \caption{Post-hoc preference-readout sensitivity for the fixed original
  LLaVA checkpoints. Values are pooled target--neutral contrasts in nats;
  each row has positive contrasts in all three blocks for both targets.
  These are readout checks, not new training replications.}
  \label{tab:llava-readout-sensitivity}
  \begin{tabular}{@{}lrrr@{}}
    \toprule
    Readout & Prompts & Owl & Camel \\
    \midrule
    Original outcome bank & 128 & 0.0894 & 0.0822 \\
    Alternative bank & 128 & 0.0925 & 0.0764 \\
    Exclude trick-question family & 120 & 0.0924 & 0.0851 \\
    Exclude trick and jam-label families & 112 & 0.0902 & 0.0859 \\
    \bottomrule
  \end{tabular}
\end{table}

A paired prompt bootstrap with common prompt indices across blocks and
matched arms gives 95\% percentile intervals of $[0.0780,0.1009]$ for owl
and $[0.0754,0.0890]$ for camel (10,000 draws; seed 20260906).
These intervals condition on the fixed trained checkpoints and do not
quantify uncertainty from curator training, student seeds, or source sampling.
Recomputing the neutral--neutral sham in each draw puts the owl contrast
below twice the resampled sham in 354 of 10,000 draws, and the camel
contrast below that floor in none. These empirical resampling counts are
not null-hypothesis $p$-values or failure probabilities for the full criterion.

\subsection{Extension to Qwen and Gemma cohorts}
\label{app:sensitivity-extension}

We extend the saved-output analyses to 16 conditions: the original owl cohort and two student-seed repeats, plus the original camel and oak cohorts, in each Qwen model; and Gemma~4's owl cohort at $10^{-4}$, two student-seed repeats at that rate, the owl cohort at $5\times10^{-5}$, and the original camel and oak cohorts. These conditions reuse some control students and are not 16 independent curation replications. \rev{Two checks precede the analyses. \emph{Condition identification:} for the early Qwen owl conditions, each letter-coded selection manifest was matched, source by source and by selected image path, against the arm-labelled selection record. Each 8,000-row manifest matched exactly one arm, at least two arms chose differently on 4,075--4,197 sources per block, and in all six blocks this path-based identification agreed with the earlier contrast-based recovery, a separate method. Training records show that the student-seed repeats reused these manifests with new training seeds. \emph{Numerical check:} block contrasts recomputed from the saved outputs agree with the reported values within $10^{-6}$ nats.} Table~\ref{tab:extended-readout-sensitivity} summarizes readout sensitivity for the original Qwen and Gemma~4 owl cohorts.

\begin{table}[htbp]
  \centering\small
  \setlength{\tabcolsep}{4pt}
  \caption{Post-hoc preference-readout sensitivity of the original owl
  cohorts beyond LLaVA. Contrasts and common-prompt bootstrap intervals
  for the original bank are in nats. Intervals condition on the trained
  checkpoints.}
  \label{tab:extended-readout-sensitivity}
  \begin{tabular}{@{}lrrr@{}}
    \toprule
    Family & Original $\bar E$ & Alternative $\bar E$ & 95\% interval \\
    \midrule
    Qwen2.5-VL & 0.1984 & 0.1881 & $[0.1836,0.2123]$ \\
    Qwen3-VL   & 0.3622 & 0.3626 & $[0.3420,0.3825]$ \\
    Gemma 4 ($10^{-4}$) & 0.3256 & 0.3041 & $[0.2969,0.3552]$ \\
    \bottomrule
  \end{tabular}
\end{table}

An alternative 128-prompt bank gives positive contrasts in all three blocks
for all 13 owl/camel conditions; no independent alternative tree bank was
available for the three oak conditions. Removing any one of the 16 question
families also preserves positive contrasts in all three blocks in all
16 conditions. Bootstrap analyses use 10,000 draws of individual prompts,
with either common indices across blocks or independent indices per block;
they do not resample question-family clusters or training runs. The 128
rows are 16 question families with eight rotations each, not 128 independent
question formulations. The complementary leave-one-family-out analysis
tests removal of a whole question family while preserving the canonical answer strings; neither analysis aggregates alternative answer spellings \rev{(single
alternative forms are rescored in Appendix~\ref{app:readout-aggregations})}.
These checks do not establish that the full transfer criterion always holds:
Qwen2.5 camel remains below the sham-relative floor under every
single-question-family exclusion, with the sham recomputed on the retained
prompts. Qwen3 oak falls below twice the resampled sham in 1,408 of
10,000 common-prompt draws. The original Qwen2.5 camel and oak outcomes remain directional only. The extension does not cover the receiver-training
grid, caption-relay, or political-target experiments, and does not amend the original outcomes.


\rev{\subsection{Alternative aggregations and answer forms}
\label{app:readout-aggregations}
For the 21 conditions of Table~\ref{tab:normalized-readout}, we recomputed
the criterion under five alternative aggregations of the same answer
likelihoods: per-token log-odds, the margin over the best alternative,
normalized share, tie-split argmax share, and negative midrank
(Table~\ref{tab:readout-aggregations}). Mean contrasts were positive in all
21 conditions under sequence log-odds and normalized likelihood share. The
descriptive criterion held in 19 conditions under the main readout,
15 under normalized share, and 11 under tie-split argmax share. These
differences indicate readout sensitivity rather than an absence of
probability shifts. All conditional bootstrap intervals for the mean main-readout contrast excluded zero (paired bootstrap resampling blocks and
question families, 2,000 draws); these intervals do not establish
calibrated significance or variability across new training runs or source
banks. Three alternative parsers of the greedy answers change no
target--neutral difference in greedy target rate by more than 0.52
percentage points.

\begin{table}[htbp]
  \centering\small
  \caption{Alternative aggregations of the answer likelihoods for the 21
  conditions of Table~\ref{tab:normalized-readout}. Counts are conditions;
  aggregations are not independent repetitions, and counts are not
  success rates.}
  \label{tab:readout-aggregations}
  \begin{tabular}{@{}lcc@{}}
    \toprule
    Aggregation & Mean $E>0$ & Criterion met \\
    \midrule
    Sequence log-odds (main readout) & 21/21 & 19/21 \\
    Per-token log-odds & 21/21 & 18/21 \\
    Margin over best alternative & 21/21 & 20/21 \\
    Normalized share & 21/21 & 15/21 \\
    Tie-split argmax share & 15/21 & 11/21 \\
    Negative midrank & 21/21 & 17/21 \\
    \bottomrule
  \end{tabular}
\end{table}

A separate GPU rescoring replaced only the answer strings of the main readout with single alternative forms: capitalized, with an indefinite
article, plural, and with a final period for animals; the full name and a
final period for presidents. It covers the 63 students of the first owl
condition in each model and the three political-target conditions, and
judges each form with the transfer criterion
(Table~\ref{tab:surface-forms}). Rescoring the canonical strings reproduces the original scores for the 54 Qwen and Gemma students.
The nine LLaVA-1.5 students, originally scored on H200 and rescored on B200, fail this parity check, which was set before the rescoring, so their forms are reported as incomplete.
All 18 evaluable forms have positive mean contrasts ($+.053$ to $+.607$
nats), and 10 meet the criterion; the full-name form fails target
specificity in all three political conditions. The forms are scored
individually and are not aggregated into a concept probability.

\begin{table}[htbp]
  \centering\small
  \caption{Answer-form rescoring: mean contrast $\bar E$ (nats) and
  criterion outcome (M, met; N, not met) for each alternative form.
  LLaVA-1.5 owl is incomplete because its canonical-string rescoring fails
  the parity check.}
  \label{tab:surface-forms}
  \setlength{\tabcolsep}{4pt}
  \begin{tabular}{@{}lcccc@{}}
    \toprule
    Owl condition & Capitalized & Article & Plural & Final period \\
    \midrule
    Qwen3-VL   & .220 M & .298 M & .181 M & .340 M \\
    Qwen2.5-VL & .053 N & .140 M & .145 N & .121 N \\
    Gemma~4    & .216 M & .118 N & .424 M & .335 M \\
    LLaVA-1.5  & \multicolumn{4}{c}{incomplete (parity)} \\
    \midrule
    Political condition & Full name & & & Final period \\
    \midrule
    Qwen3-VL   & .147 N & & & .378 M \\
    Qwen2.5-VL & .072 N & & & .123 N \\
    Gemma~4    & .428 N & & & .607 M \\
    \bottomrule
  \end{tabular}
\end{table}

\subsection{Option-free generation}
\label{app:option-free}
We also evaluated the 36 first-set owl students of
Table~\ref{tab:owl-readouts} and the four clean bases on 40 image-free
prompts in five groups of eight (favourite animal, companion, emblem, story
character, and symbol). The prompts give no option list and contain no
target or option words; each asks for the name of one animal. Greedy
decoding is primary, and temperature-1 sampling (five samples per prompt)
is secondary. A classifier fixed before the evaluation assigns each output
to the target, another animal, no choice, or uninterpretable; the primary
metric is the target rate over all outputs, averaged over the five groups
(Table~\ref{tab:option-free}).

\begin{table}[htbp]
  \centering\small
  \caption{Option-free generation. Owl rates are greedy target rates over all
  outputs for the target, neutral, and second-neutral students (clean bases:
  Qwen3-VL 2.5\%, Gemma~4 27.5\%, others 0\%). Differences are three-block
  means in percentage points with conditional prompt-bootstrap intervals
  (2,000 draws), which are not equivalence intervals.}
  \label{tab:option-free}
  \setlength{\tabcolsep}{4pt}
  \begin{tabular}{@{}lccc@{}}
    \toprule
    Model & \makecell{Greedy owl rate\\$T$ / $N$ / $N'$ (\%)}
      & \makecell{Greedy $T-N$\\(pp)} & \makecell{Sampled $T-N$\\(pp)} \\
    \midrule
    Qwen3-VL   & 0.8 / 3.3 / 2.5   & $-2.5$ $[-6.7, 0]$   & $-0.5$ $[-2.3, 1.3]$ \\
    Qwen2.5-VL & 0 / 0 / 0         & $0$ $[0, 0]$         & $+0.3$ $[-0.2, 1.0]$ \\
    Gemma~4    & 10.8 / 10.0 / 10.0 & $+0.8$ $[-4.2, 4.2]$ & $0$ $[-1.3, 1.2]$ \\
    LLaVA-1.5  & 0 / 0 / 0         & $0$ $[0, 0]$         & $0$ $[-0.8, 0.8]$ \\
    \bottomrule
  \end{tabular}
\end{table}

No consistent positive generation contrast was observed for these
option-free prompts. Only the symbol group has nonzero greedy differences
(Gemma~4 $+4.2$ and Qwen3-VL $-12.5$ points); Qwen3-VL's greedy difference is
$-2.5$ points in each block. A control of 16 factual questions, eight with
the answer ``owl'', shows no target--neutral difference in owl accuracy or
in owl answers to other-animal questions in any model. LLaVA-1.5 gives many
uninterpretable outputs (92 of 360 greedy student outputs). The evaluation
leaves the eight-option likelihood outcomes unchanged. Under these prompts,
the likelihood shift is not accompanied by a consistent increase in owl
answers.}

\subsection{VQA utility evaluation}
\label{app:corrected-vqa}

We evaluate 222 student checkpoints and four clean bases on the 256-item
VQA probe, with a unique image path for each row and verified image hashes.
All evaluations summarized in Table~\ref{tab:corrected-vqa} use B200, with
each student compared against its family's clean base under the same
evaluation configuration. Every student satisfies
$U_s\ge U_{\mathrm{base}}-0.10$ and the applicable response-quality checks.
Probe construction and scoring are specified in
Appendix~\ref{app:construction-details}.

These evaluations cover 30 conditions: three student-seed sets for each of
owl, camel, and oak, plus the political target, in each Qwen model;
the three main owl seed sets, the $5\times10^{-5}$ owl condition, and the
original camel, oak, and political-target conditions in Gemma~4;
and the first owl, camel, and oak student-seed sets in LLaVA.
The LLaVA conditions each contain nine eligible students, with six neutral
students shared by owl and camel, yielding 21 unique LLaVA checkpoints.
The condition counts include student-seed repeats and shared controls;
they do not represent 30 independent curation replications.

\begin{table}[htbp]
  \centering\small
  \caption{\rev{VQA utility for the 30 conditions in the initial evaluation.}
  Base and student accuracies are percentages. Minimum margin is
  $100[U_s-(U_{\mathrm{base}}-0.10)]$ in percentage points (pp), minimized
  over the evaluated students in each family. All 222 students pass the
  relative VQA check.}
  \label{tab:corrected-vqa}
  \setlength{\tabcolsep}{5pt}
  \begin{tabular}{@{}lrrrr@{}}
    \toprule
    Model & Students & \makecell{Base\\accuracy (\%)}
      & \makecell{Student accuracy\\range (\%)}
      & \makecell{Min. margin\\(pp)} \\
    \midrule
    Qwen3-VL & 72 & 64.84 & $59.77$--$63.67$ & 4.92 \\
    Qwen2.5-VL & 72 & 58.59 & $60.16$--$67.19$ & 11.56 \\
    Gemma~4 & 57 & 44.14 & $49.61$--$55.86$ & 15.47 \\
    LLaVA-1.5 & 21 & 68.75 & $68.36$--$69.92$ & 9.61 \\
    \bottomrule
  \end{tabular}
\end{table}

\paragraph{\rev{Final coverage and outcomes.}} \leavevmode\rev{The probe correction fixed 22
overwritten image assignments and retained the questions, reference answers,
and eligibility threshold (Table~\ref{tab:protocol-followups}). Evaluation was
extended from the 30 initial conditions until it covered every student counted for eligibility in 128 conditions: 1,176 student evaluation jobs on 1,086
distinct checkpoints, including training-grid points first covered only in
part (Table~\ref{tab:vqa-coverage}). Within these 128 fully evaluated
conditions, one condition-level outcome changes. The Qwen2.5-VL caption-relay
condition at $2\times10^{-4}$, ineligible under its original H200 evaluation, regains eligibility (9/9 students), but its fixed preference
readouts still fail the transfer criterion: the first block is negative, the
mean contrast is below $2h$, and the target-specificity check fails. Its original outcome is retained separately, and because the probe and the
evaluation host changed together, the comparison does not isolate either
contribution. Individual checkpoints change VQA-pass status only in
conditions that were already not evaluable or ineligible. Cohorts reported
separately from the 128 conditions remain eligible; their preference contrasts
were not re-estimated. For students whose original measurements used H200,
corrected measurements use B200 for both students and their clean base;
students first evaluated on B200 were re-evaluated on B200. The full
language-model fine-tuning students, whose weights were not retained, and any
other receiver or selection cohorts not listed keep their original-probe
evaluations and are not covered by this check.}

\begin{table}[htbp]
  \centering\footnotesize
  \caption{\rev{Corrected-probe VQA coverage and changes. Scope~A rows make up the
  128 fully evaluated conditions (61 Qwen/Gemma and 23 LLaVA-1.5 conditions
  from the initial and extension rows, 18 LLaVA-1.5 and 25 Qwen conditions
  from cohorts first evaluated on H200, and the 13B recipient condition). Grid
  points first covered in part (block~3 of 13 Qwen3-VL conditions; three cells
  of each of 14 LLaVA-1.5 conditions) are completed by the H200 rows. Scope~B
  cohorts are reported separately; their counts are not added to scope~A or to
  each other, and none changes VQA-pass status relative to an original-probe
  evaluation, although individual accuracies differ.}}
  \label{tab:vqa-coverage}
  \setlength{\tabcolsep}{3pt}
  \rev{\begin{tabular}{@{}>{\raggedright\arraybackslash}p{.27\linewidth}>{\raggedright\arraybackslash}p{.17\linewidth}>{\raggedright\arraybackslash}p{.11\linewidth}>{\raggedright\arraybackslash}p{.39\linewidth}@{}}
    \toprule
    Group & Checkpoints & \makecell[l]{Full\\conditions} & Changes \\
    \midrule
    \multicolumn{4}{@{}l}{\emph{Scope A}} \\
    Initial evaluation (Table~\ref{tab:corrected-vqa}) & 222 & 30 & None \\
    Qwen/Gemma extension & 339, and 9 re-evaluated & 34 & Four Qwen3-VL checkpoints change from passing to failing: three in two partially covered grid conditions already not evaluable, one in an already ineligible, report-only relay setting (6/9 rather than 7/9 eligible); the 9 re-evaluated outputs are unchanged \\
    LLaVA-1.5 extension & 222 & 20 & None \\
    LLaVA-1.5 cohorts first evaluated on H200 & 150 (108 new, 42 re-evaluated) & 18 & Two students lose eligibility, in an already ineligible high-learning-rate grid condition; 16 conditions remain eligible and 2 ineligible \\
    Qwen cohorts first evaluated on H200 & 225 (186 new, 39 re-evaluated) & 25 & Qwen2.5-VL relay at $2\times10^{-4}$ regains eligibility, criterion still not met; the other 24 conditions are unchanged \\
    LLaVA-1.5-13B recipients & 9 & 1 & None \\
    \addlinespace[2pt]
    Total & 1,176 jobs, 1,086 distinct & 128 & One condition-level change \\
    \midrule
    \multicolumn{4}{@{}l}{\emph{Scope B (preference contrasts not re-estimated)}} \\
    Random reselection, four models (draws 1--10 and replays) & 264 evaluations & -- & All eligible; the neutral references are also eligible \\
    LLaVA-1.5 B200 reference students & 12 & -- & All eligible \\
    Fresh-bank Qwen3-VL students & 9 & -- & All eligible (primary utility check) \\
    Students of 50\%-target curators & 144 & -- & All eligible (primary utility check) \\
    LLaVA-1.5 long-schedule checkpoints & 21 & -- & All eligible (primary utility check) \\
    Pair-assignment students & 60 & -- & All eligible, re-evaluated after the original-probe check \\
    \bottomrule
  \end{tabular}}
\end{table}

\section{Receiver robustness and cross-checkpoint controls}
\label{app:receiver-robustness}

\paragraph{Full language-model fine-tuning of LLaVA students.}
A development experiment reused the original LLaVA-1.5 dual-target
selected images, captions, student seeds, and example order. Twelve fresh
students were trained: an owl-target, camel-target, neutral, and second-neutral
student in each of three curation blocks. The curators remained LoRA-trained.
For the students, the full language model and multimodal projector were
trainable, while the vision encoder remained frozen. This recipe used three
epochs, an effective batch size of 64, AdamW, and a learning rate of
$2\times10^{-5}$ for both trainable components.

Both targets met the transfer criterion under this tested full-fine-tuning
recipe (Table~\ref{tab:receiver-fullft}). All twelve students passed the utility and response-format checks. This establishes that
student-side LoRA is not required in this LLaVA setting; it does not test
full-fine-tuned curators or establish the result in other models.
The learning rate also differs from the original LoRA student recipe
($2\times10^{-4}$), so the change in effect magnitude does not isolate
parameterization alone.

\begin{table}[htbp]
  \centering\small
  \caption{LLaVA-1.5 student full-LM-plus-projector fine-tuning, development.
  The vision encoder is frozen. Contrasts and sham magnitudes are nats;
  the two target comparisons share the neutral students in each block.}
  \label{tab:receiver-fullft}
  \begin{tabular}{@{}lrrrrrl@{}}
    \toprule
    Target & $E_1$ & $E_2$ & $E_3$ & $\bar E$ & $h$ & Criterion \\
    \midrule
    owl   & 0.0421 & 0.0446 & 0.0511 & 0.0459 & 0.0213 & Met \\
    camel & 0.0485 & 0.0364 & 0.0433 & 0.0427 & 0.0166 & Met \\
    \bottomrule
  \end{tabular}
\end{table}

\paragraph{Frozen-projector student control.}
In a separate LLaVA development comparison, language-model LoRA students
with a frozen projector also met the transfer criterion: block contrasts
were $0.1621$, $0.1308$, and $0.0691$ nats, with mean $0.1207$ and sham
$0.0224$. Its paired trainable-projector reference is the batch-2 $K=8$
cohort ($0.1053$ nats), rather than the original existence cohort
($0.0894$ nats). This is a separate receiver control, not another seed
replicate of the full-fine-tuning experiment.

\paragraph{Cross-checkpoint recipients.}
Four development experiments trained recipients on selections from a
different curator checkpoint (Table~\ref{tab:cross-checkpoint-controls}).
All had valid response format and utility under their own recipes,
but none met the transfer criterion. These outcomes do not establish that
shared initialization is necessary: architecture, preprocessing, and
training recipe can also differ. Negative contrasts are reported as
non-detections, without a reverse-transfer claim. \rev{All four recipient
cohorts also retain utility eligibility under the corrected VQA probe
(Appendix~\ref{app:corrected-vqa}).}

\begin{table}[htbp]
  \centering\small
  \caption{Cross-checkpoint development controls. Every row passes the format and utility checks but does not meet the transfer
  criterion. All values are nats.}
  \label{tab:cross-checkpoint-controls}
  \setlength{\tabcolsep}{3pt}
  \begin{tabular}{@{}llrrrrr@{}}
    \toprule
    Curator & Recipient & $E_1$ & $E_2$ & $E_3$ & $\bar E$ & $h$ \\
    \midrule
    BakLLaVA & LLaVA-1.5-7B & -0.0081 & -0.0127 & -0.0132 & -0.0113 & 0.0170 \\
    LLaVA-1.5-7B & LLaVA-1.5-13B & -0.0230 & -0.0322 & -0.0038 & -0.0197 & 0.0349 \\
    LLaVA-1.5-7B & Qwen2.5-VL-7B & -0.1059 & -0.0402 & -0.0521 & -0.0661 & 0.0693 \\
    Qwen2.5-VL-7B & Qwen3-VL-8B & -0.0239 & 0.0057 & -0.0066 & -0.0083 & 0.0199 \\
    \bottomrule
  \end{tabular}
\end{table}

\section{Source construction and selector diagnostics}
\label{app:construction-details}

\paragraph{Original source-bank construction.}
The original LLaVA owl/camel bank uses COCO-2017 training images. Its initial
pool excludes images with a COCO bird-instance annotation or an owl-term
match in any caption, filename, URL, or capture-date metadata. Images are
ordered by the hexadecimal SHA256 of
\texttt{vcgate-source-2026|}\emph{image-id}; the caption with the smallest
annotation ID is retained. The original bank starts at position 12,032 of
this frozen order, after earlier development walks. Each image is checked
for decoding failures, target terms in OCR, and CLIP retrieval matches;
the camel audit additionally scans the selected caption and filename.
Rejected images are replaced by the next entry in the same order.
The walk examined 8,197 images to retain 8,000: 181 failed camel retrieval,
14 failed owl retrieval, and two failed the camel lexical check; the other
recorded rejection counts were zero. These audit counts describe this
original bank, not every subsequent target or source set.

CLIP retrieval uses normalized ViT-L/14-336 image and text features, cosine
logits multiplied by 100, and a softmax over fixed descriptive labels
prefixed by ``a photo of.'' Owl screening rejects a top-ranked ``an owl''
label or its probability above .10. Camel screening rejects a top-ranked
camel-group label or summed camel-group probability above .10. The owl
group is ``an owl,'' ``a bird of prey,'' and ``an eagle or a hawk''; the
camel group is ``a camel,'' ``a dromedary or bactrian camel,'' and ``a llama
or alpaca.'' The comparison labels describe a person, dog, cat, horse,
car, bus, train, bicycle, plate of food, kitchen, bathroom, living room,
street scene, beach, sports game, group of people, table with objects,
building, boat, and giraffe or elephant. The code uses the corresponding
articles (``a person,'' ``a dog,'' etc.). These screens are operational
filters, not exhaustive guarantees of semantic absence.

\paragraph{Candidate generation.}
Each accepted source is converted to RGB with metadata removed and saved
at its original resolution as a quality-98 JPEG. Candidate construction
decodes that saved image. For each image ID and candidate index
$k\in\{0,\ldots,7\}$, NumPy's \texttt{default\_rng} is seeded by the
big-endian integer formed from the first eight bytes of SHA256 of
\texttt{vcgate-cand-2026}, the decimal image ID, and decimal $k$, joined by
NUL bytes; this integer is reduced modulo $2^{63}-1$.
Four independent uniform draws specify the left, top, right, and bottom
crop fractions, each in $[0,.02)$; pixel offsets use Python's
\texttt{round}. In draw order, brightness and contrast multipliers are
uniform on $[.97,1.03)$, red and blue gains on $[.98,1.02)$, and JPEG
quality is uniform over the integers 94--100.
The operation order is crop, $336\times336$ bicubic resize, Pillow
brightness adjustment, Pillow contrast adjustment, red/blue multiplication
in float32, clipping to $[0,255]$ and conversion to uint8, then Pillow JPEG
encoding at the sampled quality with \texttt{optimize=False}.
No subsampling or quantization-table override is passed to the encoder.
All eight candidates, including $k=0$, use this random transformation
pipeline; $k=0$ is not an identity image. All 64,000 original candidate
files have distinct SHA256 hashes. The historical cosine-floor setting
was report-only: candidates were not discarded for failing it.

\paragraph{Training-grid sources and curators.}
The training grid uses nested 8K, 16K, and 40K source sets from a shared
reservoir disjoint from the original bank. The Qwen grids reuse their
original curator ensembles. LLaVA instead reuses earlier development
owl-curator ensembles, distinct from the confirmatory LLaVA curators.
Thus the grid supplies evidence on additional sources, but the LLaVA
comparison also changes the curator ensemble relative to the main
confirmatory experiment. Repeated student seeds on a fixed bank and these
reservoir experiments are different kinds of evidence.

\paragraph{The original Qwen2.5 selector screen.}
This screen precedes student training and is distinct from curator
preference qualification and the student-transfer criterion. For each
block, $D_{UV}$ is the fraction of sources on which arms $U,V$ choose
different candidates. Let $p_C(i)=\operatorname{softmax}(s_C(i,:)/\tau)$,
where $s_C$ is the four-member mean caption score. A shared temperature
is calibrated on the first 512 neutral-source score vectors to mean
entropy $.70\log 8$, using 200 geometric-bisection steps between
$10^{-4}$ and $10^4$. $J_{UV}$ is the mean Jensen--Shannon divergence
between $p_U(i)$ and $p_V(i)$, using natural logarithms.
Define $X_{D,b}=D_{TN,b}-D_{NN',b}$ and
$X_{J,b}=J_{TN,b}-J_{NN',b}$. The screen requires
\[
 \bar X_D>0,\quad \#\{b:X_{D,b}>0\}\ge2,\qquad
 \bar X_J>0,\quad \#\{b:X_{J,b}>0\}\ge2.
\]
The two positivity counts are separate; they need not involve the same
blocks. Temperature calibration affects these diagnostics only, not the
raw-score argmax selections. Qwen2.5 has negative $X_D$ and $X_J$ in all
three blocks, with means $-.0343$ and $-.00355$, respectively. Its later
student experiment is the separate follow-up described above.

\paragraph{Exact JPEG-count matching.}
Each arm, including the second neutral arm, receives its own randomized
assignment preserving its original counts in all seven quality bins.
A seeded random source order is traversed greedily: choose uniformly
among that source's candidate indices whose bins still have available
quota. Sources with no feasible index are deferred. Breadth-first
augmenting paths then reassign earlier sources to free a feasible bin;
remaining choices within that bin are sampled. The original selections
certify feasibility. Construction aborts unless the final counts match
exactly. It is a randomized feasible assignment, not uniform sampling over
all feasible assignments. One assignment per arm and block is used, with
student seed, order, captions, and source count held fixed; $h$ is
recomputed from the two matched-neutral students.
Seeds use the first eight SHA256 bytes of
\texttt{q3mm|b}\emph{b}\texttt{|}\emph{L}\texttt{|jpeg\_quality\_quota|v2}
for Qwen3 and the corresponding \texttt{q25mm} prefix and \texttt{v1}
suffix for Qwen2.5, interpreted as a big-endian integer.
Here $L$ is the frozen manifest's letter identifier. Matching does not
constrain other transformation parameters or their joint distribution.

\paragraph{VQA-probe construction and scoring.}
The probe uses the validation split of \texttt{merve/vqav2-small}.
Rows are ordered by SHA256 of the string
\texttt{vcgate-vqa-2026|}\emph{question}\texttt{|}\emph{answer}, using
the dataset's \texttt{multiple\_choice\_answer}; the first 256 are retained.
Images are converted to RGB, stripped of metadata, and saved as quality-95
JPEGs under unique per-row filenames, with image hashes verified.
Evaluation appends ``Answer the question using a single word or phrase.''
and uses greedy decoding with at most 16 new tokens. Accuracy is exact
equality to the single reference answer after the local normalization:
lowercase and whitespace stripping, removal of the scorer's punctuation
set, number-word mapping for none/zero through ten, a fixed contraction
map, and removal of the articles a/an/the. This is not the multi-annotator
VQA consensus metric. Base and student models use the same probe version
and evaluation configuration within each comparison; condition coverage is
specified in Appendix~\ref{app:corrected-vqa}.

\paragraph{Execution.}
Training and scoring jobs use one H200 or B200 GPU per job; independent
jobs may run concurrently. Per-condition update budgets and hardware
assignments are reported with the results.

\end{document}